# LLmFPCA-detect: LLM-powered Multivariate Functional PCA for Anomaly Detection in Sparse Longitudinal Texts

Prasanjit Dubey[1,*]   Aritra Guha[2]   Zhengyi Zhou[2]   Qiong Wu[2]

Xiaoming Huo[1]   Paromita Dubey[3]

[1]H. Milton Stewart School of Industrial and Systems Engineering,

Georgia Institute of Technology

[2]AT&T Chief Data Office

[3]Department of Data Sciences and Operations, Marshall School of Business,

University of Southern California

## Abstract

Sparse longitudinal (SL) textual data arises when individuals generate text repeatedly over time (e.g., customer reviews, occasional social media posts, electronic medical records across visits), but the frequency and timing of observations vary across individuals. These complex textual data sets have immense potential to inform future policy and targeted recommendations. However, because SL text data lack dedicated methods and are noisy, heterogeneous, and prone to anomalies, detecting and inferring key patterns is challenging. We introduce LLmFPCA-detect, a flexible framework that pairs LLM-based text embeddings with functional data analysis to detect clusters and infer anomalies in large SL text datasets. First, LLmFPCA-detect embeds each piece of text into an application-specific numeric space using LLM prompts. Sparse multivariate functional principal component analysis (mFPCA) conducted in the numeric space forms the workhorse to recover primary population characteristics, and produces subject-level scores which, together with baseline static covariates, facilitate data segmentation, unsupervised anomaly detection and inference, and enable other downstream tasks. In particular, we leverage LLMs to perform dynamic keyword profiling guided by the data segments and anomalies discovered by LLmFPCA-detect, and we show that cluster-specific functional PC scores from LLmFPCA-detect, used as features in existing pipelines, help boost prediction performance. We support the stability of LLmFPCA-detect with experiments and evaluate it on two different applications using public datasets, Amazon customer-review trajectories, and Wikipedia talk-page comment streams, demonstrating utility across domains and outperforming state-of-the-art baselines.

*Corresponding author. 755 Ferst Dr NW, Atlanta, GA 30332, USA. Email: pdubey31@gatech.edu

# 1 Introduction

In modern machine learning, it is common to encounter datasets comprising of $N$ subjects, where each subject $i$ is associated with a sequence of textual observations $\{K_i(T_{i1}), K_i(T_{i2}), \ldots, K_i(T_{iN_i})\}$ recorded at sparse and irregular time points $\{T_{i1}, T_{i2}, \ldots, T_{iN_i}\} \subset \mathbb{R}$. Despite LLMs having spurred many advancements in analysis of text data, current methods are not well adapted to sparse longitudinal (SL) designs—time-evolving texts observed at irregular, subject-specific times—so these are frequently discarded or collapsed across time, ignoring the dynamic patterns in the texts. In this paper, we propose a novel framework for the analysis of SL text data that yields representations suitable for straightforward integration into unsupervised and supervised learning pipelines. The proposed methodology is applicable to a wide range of domains that generate SL text data, such as, electronic medical records in healthcare [Ford et al., 2016], consumer interactions through service channels in business [Cavique et al., 2022], activity logs from online learning platforms in education [Yang and Kang, 2020], user posts and comments on social media Hutto et al. [2013], Valdez et al. [2020], Kelley and Gillan [2022] and many more.

A major challenge with SL text datasets is that observations are unstructured and noisy, heterogeneous across subjects, and may contain outliers. The first step in making such data amenable for downstream supervised or unsupervised learning tasks, including prediction and inference, is to extract parsimonious feature representations of the longitudinal texts that capture the leading modes of variation. In this work, we propose LLmFPCA-detect, which starts from noisy SL texts and produces learned representations, accounting for heterogeneity and providing type-I-error–controlled outlier screening. LLmFPCA-detect begins by embedding text into an application specific numeric space using LLMs. In this numeric space, sparse multivariate functional principal component analysis(mFPCA) Happ and Greven [2018], Yao et al. [2005] is used to model the longitudinal text embeddings as noisy observations of an underlying smooth trajectory. The method first clusters the preliminary FPC scores, augmented with baseline subject-level covariates, and then screens for outliers; a novel calibration step yields the final set of anomalies with statistical significance guarantees. We illustrate this new approach on two datasets: the Amazon review corpus and the Wikipedia talk- page comment stream, where LLmFPCA-detect reveals insightful findings from SL text data.

**Related Works** *Modeling SL data* Beginning with the seminal parametric random-effects formulation Laird and Ware [1982], the field of longitudinal data analysis has undergone extensive development over the decades; see Verbeke et al. [2014] for a review on multivariate longitudinal data analysis. Functional data analysis (FDA) provides a nonparametric framework for SL data—via principal components through conditional expectation Yao et al. [2005], Happ and Greven [2018]—to predict subject-specific smooth trajectories even from one or a few observations. While this line of work has expanded to include dynamic Hao et al. [2024], Zhou and Mueller [2024] and covariate-dependent Kim et al. [2023] extensions, and has led to methods for clustering and unsupervised anomaly detection Schmutz et al. [2020], Wu et al. [2023], Castrillón-Candás and Kon [2022], and supervised tasks such as regression and classification Müller [2005], none of these methods extend directly to heterogeneous, complex SL text data paired with baseline covariates and containing outliers.

*Text time series versus SL texts* An SL design differs from a time series; instead of a single, regularly spaced sequence of observations, it comprises many subjects, each with its own trajectory recorded at irregular, subject-specific times where per-subject sampling is sparse, and between-subject heterogeneity could be substantial. While text time-series modeling has advanced considerably O'Connor et al. [2010], Blei and Lafferty [2006], Wang and McCallum [2006], Bamler and Mandt [2017], Dodds et al. [2011], Griffiths and Steyvers [2004], Yurochkin et al. [2019], these approaches rely on dense, uniformly spaced observations and are not suited to SL texts.

*Anomaly detection* Text clustering and anomaly detection are central NLP tasks, used to flag harmful content, phishing, and spam. With pretrained language models (e.g., BERT Devlin et al. [2019], RoBERTa Liu et al. [2019], GPT Brown et al. [2020]), embedding-based detectors have proliferated alongside other approaches Yin and Wang [2016], Cao et al. [2025], Ruff et al. [2019], Subakti et al. [2022], Dhillon and Modha [2001], Liu et al. [2008], Kannan et al. [2017]. Yet three limitations persist: (i) most methods lack type-I error control for flagged anomalies; (ii) time series anomaly detectors Blázquez-García et al. [2021], Zamanzadeh Darban et al. [2024], Xu et al. [2022] can be adapted to unstructured texts via embeddings, but only assuming dense, regularly sampled streams; and (iii) these methods do not support SL designs with subject-specific, irregular observation times and evolving trajectories, hence missing on the individual level dynamic trends in the anomalies. Functional data analysis methods for SL anomaly detection exist [Sun and Genton, 2011, Dai and Genton, 2018, Hubert et al., 2015, Gervini, 2009], but they operate on structured numeric functions rather than unstructured text and likewise lack formal false-positive guarantees. As a result, there is no end-to-end solution that transforms SL texts into trajectory-aware feature representations and detects anomalies with explicit type-I error control.

**Our Contributions** We introduce LLmFPCA-detect, a novel framework that combines LLM-based embeddings with sparse mFPCA to enable covariate-informed data segmentation and type-I error controlled anomaly detection in sparsely observed, longitudinal, heterogeneous text data, yielding feature representations suitable for incorporating SL texts in a wide range of downstream tasks. LLmFPCA-detect is broadly applicable to settings involving subjects with time-stamped text records that arrive irregularly over time. While we focus on sparsely sampled scenarios, the methodology can be readily adapted to densely observed data. We demonstrate the effectiveness of LLmFPCA-detect through its application to the Amazon Reviews dataset (Amazon data) and the Wikipedia talk-page comment streams (Wiki data). The key components of the framework, as illustrated in Figure 1, are:

1. **Representation** We derive domain-appropriate LLM embeddings for each time-stamped text. For the Amazon Reviews dataset, we embed the texts using emotion scores based on Plutchik's Wheel of EmotionsPlutchik [1980], which identifies eight primary emotions as the foundation for all others. For the Wikipedia request–comment stream, we obtain toxicity and aggression scores using GPT for each comment to compare against findings from human-annotated scores.

2. **Learning trajectory representations and detection with guarantees** The numeric trajectories form multivariate SL data, which are processed using the mFPCA pipeline to obtain multivariate functional principal component (mFPC) scores. These scores, combined with baseline covariates, are used for covariate-informed clustering. Anomalies are then detected in an unsupervised manner by: i) screening points in the tails of the cluster-specific mFPC score distributions, and ii) statistically testing the screened points while controlling for multiple comparisons. The identified anomalies are further analyzed to localize time window specific deviations in the population.

3. **Interpretability and insights** We use LLMs to extract keywords from texts associated with each cluster and flagged window, revealing dynamic, human-interpretable signals that explain why the flagged discovery matters.

**Organization** The rest of the paper is organized as follows. Section 2 provides the motivation for the clustering and anomaly detection steps of LLmFPCA-detect. Section 3 outlines the methods, estimation procedures, and algorithms that make up the different steps in LLmFPCA-detect. Sections 4.1 and 4.2 demonstrates the application of LLmFPCA-detect to customer journey data from Amazon reviews and to Wikipedia request–comment streams, illustrating its cross-domain applicability. Additional details and experiments are provided in the Appendix.

## 2 Motivation and Framework

In this section, we present the foundational framework underlying LLmFPCA-detect.

**Multivariate functional data representation** For each subject $i = 1, \ldots, N$, a random function $\boldsymbol{X}_i \in L^2(\mathcal{T})^p$ is observed on a discrete, potentially irregular and sparse time grid $\{T_{ij}\}_{j,i=1}^{N_i,N}$ along with baseline covariates $\boldsymbol{Z}_i \in \mathbb{R}^q$ where $(\boldsymbol{X}, \boldsymbol{Z}) \sim \mathbb{P}$, with $\mathbb{P}$ being the joint distribution of $(\boldsymbol{X}, \boldsymbol{Z})$. The population mean function is defined as $\boldsymbol{\mu}(t) = \mathbb{E}(\boldsymbol{X}(t))$, and the covariance surface for $s, t \in \mathcal{T}$ is given by $\mathbb{C}(s,t) = \mathbb{E}\{(\boldsymbol{X}(s) - \boldsymbol{\mu}(s)) \otimes (\boldsymbol{X}(t) - \boldsymbol{\mu}(t))\}$ with entries $\mathbb{C}_{ij}(s,t) = \text{Cov}(X^{(i)}(s), X^{(j)}(t))$ is assumed to satisfy the conditions of Proposition 2 in Happ and Greven [2018]. Then, $\boldsymbol{X}$ admits a multivariate Karhunen–Loève expansion (Propositions 3 and 4 in Happ and Greven [2018])

$$\boldsymbol{X}(t) = \boldsymbol{\mu}(t) + \sum_{j=1}^{\infty} \rho_m \boldsymbol{\psi}_m(t)$$

where $\rho_m = \langle \boldsymbol{X}(t) - \boldsymbol{\mu}(t), \boldsymbol{\psi}_m(t) \rangle$ with $\text{Cov}(\rho_m, \rho_n) = \lambda_m \mathbb{I}\{m = n\}$, and $\lambda_1 \geq \lambda_2 \geq \cdots \geq 0$ are the eigenvalues of the covariance operator associated with $\mathbb{C}$. The corresponding eigenfunctions $\boldsymbol{\psi}_m$, $m \in \mathbb{N}$ serve as the multivariate functional principal components, with $\rho_m$ being the associated mFPC scores. If $\boldsymbol{X}$ admits a finite expansion with $M$ principal components, Proposition 5 in Happ and Greven [2018] establishes how mFPCA of $\boldsymbol{X}$ relates to univariate functional principal component analysis (uFPCA) of each component $X^{(d)}(\cdot) \in L^2(\mathcal{T})$ for $d = 1, \ldots, p$.

**Data heterogeneity and anomalies** Suppose the trajectories $\{\boldsymbol{X}_i\}_{i=1}^{N}$ belong to $K$ distinct clusters, denoted by $\mathcal{C}_1, \ldots, \mathcal{C}_K$, with $\bigcup_k \mathcal{C}_k = \{1, \ldots, n\}$ and $\mathcal{C}_k \cap \mathcal{C}_j = \emptyset$ for $j \neq k$. Observations in $\mathcal{C}_k$ are generated according to the distribution $\mathbb{P}_k$, yielding the overall mixture $\mathbb{P} = \sum_{k=1}^{K} \pi_k \mathbb{P}_k$ with $(\pi_1, \ldots, \pi_K)$ denoting cluster proportions. For $i \in \mathcal{C}_k$, assume that $\boldsymbol{X}_i$ admits a finite multivariate Karhunen–Loève expansion $\boldsymbol{X}_i(t) = \boldsymbol{\mu}_k(t) + \sum_{m=1}^{M} \rho_{im} \boldsymbol{\psi}_m(t), \quad i \in \mathcal{C}_k$, where $\boldsymbol{\mu}_k \in L^2(\mathcal{T})^p$ is the cluster-specific mean function, and $\boldsymbol{\psi}_m \in L^2(\mathcal{T})^p$ are shared eigenfunctions across clusters. To incorporate pos-

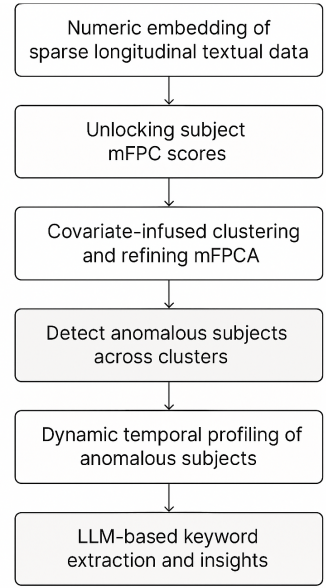

Figure 1: Proposed framework.

sible measurement errors and anomalies, we observe

$$\boldsymbol{Y}_i(t) = \boldsymbol{X}_i(t) + \boldsymbol{\eta}_i(t) + \boldsymbol{a}_i(t),$$

where $\boldsymbol{\eta}_i, \boldsymbol{a}_i \in L^2(\mathcal{T})^p$ capture the measurement errors and anomalies respectively. These are assumed to be jointly independent of $\boldsymbol{X}_i$, $i = 1, \ldots, n$, with $\mathbb{E}(\boldsymbol{\eta}_i(t)) \equiv 0$ for all $t \in \mathcal{T}$, and $\text{Cov}(\eta^{(j)}(s), \eta^{(k)}(t)) = \sigma_\eta^2 \mathbb{I}_{s=t}$ for all $j, k \in \{1, \ldots, p\}$. The term $\boldsymbol{a}_i \equiv \boldsymbol{0}$ almost surely for all $i \in \mathcal{A}_0^C$, where $\mathcal{A}_0 \subset \{1, \ldots, N\}$ denotes the set of anomalous subjects. For each $i \in \mathcal{A}_0$, we assume $\boldsymbol{a}_i(t) \neq \boldsymbol{0}$ for some $t \in \mathcal{T}_0 \subset \mathcal{T}$ almost surely. We employ trimmed $k$-means to recover the clusters accurately despite being contaminated with outliers; for details on cluster recovery see Section D.1 in Appendix D.

**Calibrating the anomalies** After the clusters are recovered, the anomalous observations in $\mathcal{A}_0$ are assigned to one of the clusters $\mathcal{C}_1, \ldots, \mathcal{C}_K$. To detect $\mathcal{A}_0$ in an unsupervised manner, we perform a screening step within each cluster by examining the tails of the FPC score distribution, approximating $\mathcal{C}_k \cap \mathcal{A}_0$ by $\mathcal{A}_0^{k,\epsilon} \subset \mathcal{C}_k$ (see Appendix E).

The distribution of FPC scores in the clean subset $\mathcal{C}_k \cap \mathcal{A}_0^C$ is then used to recover $\mathcal{C}_k \cap \mathcal{A}_0$ with confidence. In practice, each cluster $\mathcal{C}_k$ is randomly split into two subsets, and the non-screened portion is used to calibrate the anomaly detection procedure; see Theorem E.1 in Appendix E for theoretical guarantees. Finally, based on the detected anomalous set $\mathcal{A}_0$, we analyze the corresponding keywords across different time windows.

The foregoing framework outlines a pipeline for obtaining cluster-specific feature representations and type-I controlled anomaly detection in fully observed multivariate functional trajectories with possible measurement errors. In SL settings, each subject is observed at random time points $T_{ij}$ for $i = 1, \ldots, N$, $j = 1, \ldots, N_i$, with $T_{ij} \in \mathcal{T}$. These time points $T_{i1}, \ldots, T_{iN_i}$ are assumed i.i.d. and independent of $\boldsymbol{X}_i$ and $\boldsymbol{\eta}_i$ for all $i$. The number of measurements $N_i$ is random, reflecting sparse and irregular designs, and $N_i$, for $i = 1, \ldots, N$, are assumed i.i.d. and independent of all other random elements.

In practice, we observe $Y_i(T_{ij})$, $j = 1, \ldots, N_i$, $i = 1, \ldots, N$, and all relevant quantities must be estimated from these noisy observations. Section 3 outlines the estimation details and algorithms for this pipeline, including steps for incorporating the underlying textual data.

## 3 Methods: Pipeline and Estimation

**From SL Texts to Numeric Embeddings**  The first step maps each time-stamped text $K_i(T_{ij})$ to a $p$-dimensional vector via a fixed embedding

$$\Phi: \mathcal{X} \longrightarrow \mathbb{R}^p, \qquad \boldsymbol{Y}_i(T_{ij}) = \Phi\big(K_i(T_{ij})\big), \tag{3.1}$$

where $\Phi$, is implemented via LLM prompting, held constant across subjects, and deterministic (the same text yields the same vector). For subject $i$ this yields the multivariate trajectory $\{\boldsymbol{Y}_i(T_{ij})\}_{j=1}^{N_i}$, whose coordinates are modeled jointly using mFPCA (e.g. Plutchik emotion embeddings for Amazon reviews; see Sections C and 4.1). Each subject also has baseline, time-invariant covariates $\boldsymbol{Z}_i \in \mathbb{R}^q$ (e.g. average rating, review length, engagement duration).

---
**Algorithm 1** Multivariate Functional Principal Component Analysis (mFPCA)

---
**Input:** SL data: $\{\boldsymbol{Y}_i(T_{ij})\}_{j=1}^{N_i}$ for $i=1,\ldots,N$.
1: $(\{\hat{\xi}_{ik}^{(d)}\}_{i=1,k=1}^{N,K_d}, \hat{\mu}^{(d)}(t), \{\hat{\phi}_k^{(d)}(t)\}_{k=1}^{K_d}) \leftarrow \texttt{uFPCA}(\{(T_{ij}, Y_i^{(d)}(T_{ij}))\}_{i,j})$ for each dimension $d = 1,\ldots,p$.    ▷ Algorithm 4; only scores are used below
2: $\hat{\boldsymbol{\Xi}}_i \leftarrow (\hat{\xi}_{i1}^{(1)}, \ldots, \hat{\xi}_{iK_1}^{(1)}, \ldots, \hat{\xi}_{i1}^{(p)}, \ldots, \hat{\xi}_{iK_p}^{(p)})$, $i=1,\ldots,N$    ▷ Stack univariate FPC scores
3: Define matrix $\hat{\boldsymbol{\Xi}} \in \mathbb{R}^{N \times M}$ with rows $\hat{\boldsymbol{\Xi}}_i$ where $M = \sum_{d=1}^p K_d$.
4: $\hat{\mathbf{C}}_\Xi \leftarrow \frac{1}{N-1}\hat{\boldsymbol{\Xi}}^\top \hat{\boldsymbol{\Xi}}$.    ▷ Compute covariance matrix
5: Perform eigen-decomposition of $\hat{\mathbf{C}}_\Xi$ to obtain eigenvalues $\{\hat{\lambda}_m\}_{m=1}^M$ and eigenvectors $\{\hat{\boldsymbol{v}}_m\}_{m=1}^M$.
6: $\hat{\psi}_m^{(d)}(t) \leftarrow \sum_{k=1}^{K_d} \hat{v}_{m,k}^{(d)} \hat{\phi}_k^{(d)}(t)$, $d=1,\ldots,p$ and $m=1,\ldots,M$.    ▷ Multivariate eigenfunctions
7: $\hat{\rho}_{im} \leftarrow \hat{\boldsymbol{\Xi}}_i^\top \hat{\boldsymbol{v}}_m$ for $i=1,\ldots,N$ and $m=1,\ldots,M$    ▷ Compute mFPC scores

**Output:** Tuple of estimated mFPC scores, eigenfunctions and mean curves: $\{\hat{\rho}_{im}, \hat{\boldsymbol{\psi}}_m, \hat{\boldsymbol{\mu}}\}_{i,m=1}^{N,M}$.

---

**Dynamic Trajectory Representations using mFPCA**  Algorithm 1 details the estimation steps of the mFPCA setup outlined in Section 2. Starting from $\{\boldsymbol{Y}_i(T_{ij})\}_{j=1}^{N_i}$, we estimate the mFPC scores $\hat{\rho}_{im}$ by building on the univariate functional principal component analysis (uFPCA) of each $\{Y_i^{(d)}(T_{ij})\}_{j,i=1}^{N_i,N}$ for $d=1,\ldots,p$. The algorithm follows the approach in Happ and Greven [2018], using estimated quantities from uFPCA including the mean functions $\hat{\mu}^{(d)}(t)$, eigenfunctions $\hat{\phi}^{(d)}(t)$ and univariate FPC scores $\hat{\xi}_{ik}^{(d)}$; for details see Algorithm 4 in Section C.2 and Yao et al. [2005].

---

**Algorithm 2** Detecting anomalous subjects within a cluster $\hat{\mathcal{C}}$

**Input:** Subject cluster $\hat{\mathcal{C}}$; data $\{\boldsymbol{Y}_i(T_{ij}) : i \in \hat{\mathcal{C}}\}$; significance levels $\alpha_1, \alpha$ (where $\alpha_1 > \alpha$).

1: Obtain mFPC scores $\{\hat{\rho}_{im}^{\hat{\mathcal{C}}}\}_{i \in \hat{\mathcal{C}}, m=1,\ldots,B}$ corresponding to the top $B$ cluster-specific mFPC components using Algorithm 6 applied to $\{\boldsymbol{Y}_i(T_{ij}) : i \in \hat{\mathcal{C}}\}$.
   ▷ $B$: number of top mFPC components based on prop. of variance explained
2: Randomly partition $\hat{\mathcal{C}}$ into disjoint sets $I_1, I_2$ of equal size.
3: $(G_1, G_1^c) \leftarrow$ ScreenPotentialOutliers$(I_1, \{\hat{\rho}_{im}^{\hat{\mathcal{C}}} : j \in I_1\}, B, \alpha_1)$.   ▷ Algorithm 7
4: $(G_2, G_2^c) \leftarrow$ ScreenPotentialOutliers$(I_2, \{\hat{\rho}_{im}^{\hat{\mathcal{C}}} : j \in I_2\}, B, \alpha_1)$.
5: Initialize $\mathcal{A}^{(1)} \leftarrow \emptyset$.   ▷ Set of confirmed outliers for cluster $\hat{\mathcal{C}}$
6: $\mathcal{A}_{G_1}^{(1)} \leftarrow$ ConfirmAnomalies$(G_1, G_2^c, \{\hat{\rho}_{im}^{\hat{\mathcal{C}}} : j \in G_1 \cup G_2^c\}, B, \alpha)$.   ▷ Algorithm 8
7: $\mathcal{A}_{G_2}^{(1)} \leftarrow$ ConfirmAnomalies$(G_2, G_1^c, \{\hat{\rho}_{im}^{\hat{\mathcal{C}}} : j \in G_2 \cup G_1^c\}, B, \alpha)$.
8: $\mathcal{A}^{(1)} \leftarrow \mathcal{A}_{G_1}^{(1)} \cup \mathcal{A}_{G_2}^{(1)}$.

**Output:** Set of confirmed anomalous subjects $\mathcal{A}^{(1)} = \{(i, S_i) : i \in \hat{\mathcal{C}} \text{ is an outlier}, S_i \neq \emptyset\}$.

---

**Clustering and Anomaly Detection using mFPC Scores and Covariates** We segment subjects by clustering their estimated mFPC scores jointly with static covariates (Algorithm 5, Appendix D). For each estimated cluster $\hat{\mathcal{C}}_k$ we re-fit mFPCA using only its members (Algorithm 1; Algorithm 6), yielding cluster-specific means $\hat{\boldsymbol{\mu}}_k(t)$, eigenfunctions $\hat{\boldsymbol{\psi}}_m^k(t)$, updated scores $\hat{\rho}_{im}^k$ and and reconstructed trajectories (Equation (D.4)).

---

**Algorithm 3** Dynamic temporal profiling of anomalous subjects

**Input:** Type 1 anomalies $\mathcal{A}^{(1)}$ (from Alg. 2 for cluster $\hat{\mathcal{C}}$); data $\{\boldsymbol{Y}_j(T_{jk}) : j \in \hat{\mathcal{C}}\}$; cluster means $\{\hat{\mu}_{\hat{\mathcal{C}}}^{(d)}(t)\}$ (from Alg. 6); Clean held-out sets $G_1^c, G_2^c$ & split info $I_1, I_2$ for $\hat{\mathcal{C}}$ (from Alg. 2); time windows $\{(a_w, b_w]\}_{w=1}^W$; significance level $\alpha$.

1: $(\{\bar{\boldsymbol{\mu}}_{\hat{\mathcal{C}}}^{(w)}\}_{w=1}^W, \{D_j^{(w)}\}_{j \in G_1^c \cup G_2^c, w=1,\ldots,W}) \leftarrow$ ComputeWindowDeviations$(\{\boldsymbol{Y}_j(T_{jk}) : j \in G_1^c \cup G_2^c\}, \{\hat{\mu}_{\hat{\mathcal{C}}}^{(d)}(t)\}, \{(a_w, b_w]\}_{w=1}^W)$.   ▷ Compute scores for clean held-out set Alg. 9
2: $\mathcal{A}^{(2)} \leftarrow$ IdentifyAnomalousWindows$(\mathcal{A}^{(1)}, \{\boldsymbol{Y}_i(T_{ij}) : i \text{ s.t. } (i, \_) \in \mathcal{A}^{(1)}\},$
      $\{\bar{\boldsymbol{\mu}}_{\hat{\mathcal{C}}}^{(w)}\}, \{D_j^{(w)}\}, I_1, I_2, G_1^c, G_2^c,$
      $\{(a_w, b_w]\}_{w=1}^W, \alpha)$.   ▷ Identify anomalous windows for subjects (Alg. 10)

**Output:** Set of subject-indexed anomalous temporal windows $\mathcal{A}^{(2)} = \{(i, \mathcal{W}_i) : i \in \mathcal{A}^{(1)}, \mathcal{W}_i \neq \emptyset\}$.

---

Globally anomalous subjects will still be assigned to one of the $K$ clusters unless explicitly

screened—a difficult task in heterogeneous data. To detect such cases post-assignment, we apply Algorithm 2 (with Algorithms 7 and 8; Appendix E). The procedure tests whether a subject's multivariate FPC scores deviate from the typical pattern of its assigned cluster $\hat{\mathcal{C}}$, using sample splitting and data-driven calibration to control multiplicity across principal components. It outputs flagged subjects $\mathcal{A}^{(1)} = (i, S_i)$, where $S_i$ records the outlying FPC directions—information that then guides localized anomaly analysis (Algorithm 3).

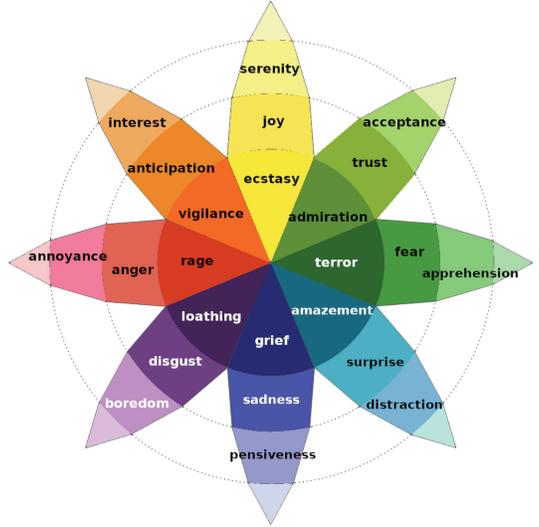

Figure 2: Plutchik's wheel of emotions.

Subjects flagged by Algorithm 2 (set $\mathcal{A}^{(1)}$) may be anomalous only over portions of their trajectories. Algorithm 3 localizes these periods by comparing each subject's raw segments to the cluster mean, with data-driven calibration (Algorithm 9); implementation details are in Appendix E (Algorithms 9, 10). The output is $\mathcal{A}^{(2)} = (i, \mathcal{W}_i)$, where $\mathcal{W}_i$ denotes the time windows in which subject $i$'s trajectory departs from a clean cohort within that window. This step pinpoints atypical intervals and enables per-window anomaly flags, which feed into the final dynamic keyword profiling stage.

**Dynamic Keyword Profiling** Finally, we describe intent extraction from anomalous reviews. For each subject $i$, let $S_i$, be the anomalous reviews. Challenges include lexical variation for similar semantics, shared stylistic drift across users, and scalability for large number of anomalous reviews. We maintain a time-ordered intent list $I_i^{(t-)}$ from reviews before time $t$. At time $t$, an LLM receives $I_i^{(t-)}$, top global intents observed before $t$, and the current review, and either matches an existing intent or proposes a new one. Full details appear in Algorithm 11 (Appendix F).

# 4 Real Data Applications

## 4.1 Modeling Dynamic Emotions in Amazon Customer Reviews

We use the Amazon Reviews corpus Hou et al. [2024], which includes 1,946 users and 22,032 reviews over five years, focusing on Automobile for the main analysis; Beauty & Personal Care and Sports & Outdoors supply user-level covariates (e.g., cross-category purchase share). Each review includes a user ID, timestamp, product title, text, and a 1–5 rating, with users posting over multiple years.

**Emotion embedding for text transcripts** Plutchik's wheel of emotions provides a structured framework for mapping emotional states along opposing pairs, capturing both intensity and polarity (see Fig. 2 Semeraro et al. [2021]). We convert each transcript into four real-valued scores—joy–sadness, trust–disgust, fear–anger, and surprise–anticipation—on a continuous $[-1, 1]$ scale, where $-1$ and $1$ denote the extremes of each pole (e.g., grief vs. ecstasy), and intermediate values encode moderate intensity. A zero-shot GPT-3.5-Turbo prompt returns one scalar per axis (details and validation in Section C.1). Stacking these over time yields a 4-D timestamped embedding per subject, which serves as the input to LLmFPCA-detect for mFPCA and the subsequent steps.

| Rating | Review | Joy–Sadness | Trust–Disgust | Fear–Anger | Surprise–Anticipation |
|---|---|---|---|---|---|
| 5 | I use this great oil in all of my 150cc Scooters (was told to by a Scooter mechanic) and I've never had an engine problem. But this price is thru the roof, $17.50 for a single quart is STUPID...wally world sells it for $4.99...but its kinda funny that all of Amazon's oils are priced thru the roof | -0.8 | -0.6 | -0.8 | **-1** |
| 1 | Received this today and went to put it on my 3/8 extension for an oil filter change. The machining is pretty, but measurements are so poor I cannot get it on the extension to use. Absolute junk! I should have paid more attention to the negative review. | -0.77 | **-0.75** | -0.5 | -0.7 |

Table 1: Amazon customer reviews with emotion scores across four Plutchik dimensions.

Table 1 illustrates how emotion embeddings reveal customer pain points that are not captured by 5-star ratings alone. In the first example, a 5-star review shows strong sadness (–0.8), disgust (–0.6), anger (–0.8), and surprise (–1), indicating frustration with pricing despite overall satisfaction. The third example, also rated 1 star, shows high sadness (–0.77), disgust (–0.75), and surprise (–0.7), pointing to severe frustration over usability issues.

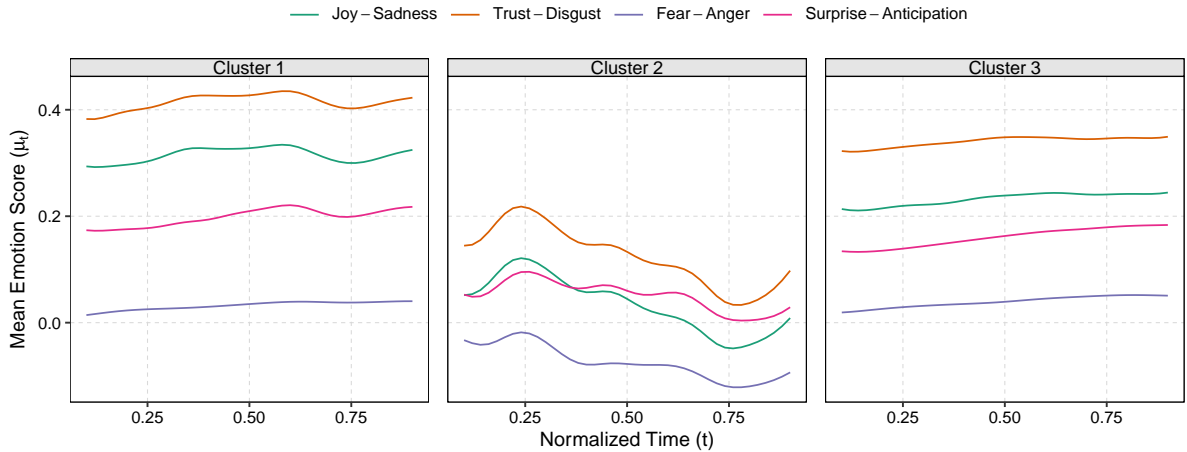

Figure 3: Mean emotion trajectories across the three user clusters. Curves represent mean scores for the Joy–Sadness, Trust–Disgust, Fear–Anger, and Surprise–Anticipation emotion dimensions.

**Emotion mFPCA scores (Algorithm 1) improve predictive power over product ratings**
We test whether review text improves forecasting of adverse outcomes (e.g., sudden rating drops) in Amazon Reviews. A "rating drop" is defined as the extreme percentile of each user's maximum gap between consecutive ratings. We compare two optimally tuned random-forest models on a class-balanced sample with identical baseline covariates—cluster labels from Algorithms 1–5 and purchase mix across categories. Model A summarizes past behavior by the mean Automobile rating; Model B replaces that single aggregate with emotion mFPC scores, capturing time-varying textual signals. On the test set, Model A: accuracy 0.542, precision 0.538, recall 0.596, F1 0.565, ROC–AUC 0.534. Model B improves all metrics—accuracy 0.609 (+**12.4%**), precision 0.610 (+**13.4%**), recall 0.603 (+**1.2%**), F1 0.606 (+**7.3%**), ROC–AUC 0.645 (+**20.8%**)–showing that compact emotion-trajectory features capture predictive signal beyond coarse star-rating averages.

**Clustering dynamics and case studies** Figure 3 plots mean emotion trajectories for the three clusters from purchasing proportions in Automobiles, Beauty & Personal Care, and Sports & Outdoors). Cluster 1 has the highest baseline across emotions–consistently stronger affect. Cluster 3 follows a similar temporal shape but is uniformly lower (milder affect). Cluster 2 departs most, with elevated sadness and anger, indicating sharper pain points. Because anomalies are scored relative to each cluster's mean, even upward shifts in positive emotion within Cluster 2 can register as anomalous. Section D.3 of the Appendix reports bootstrap analysis confirming cluster

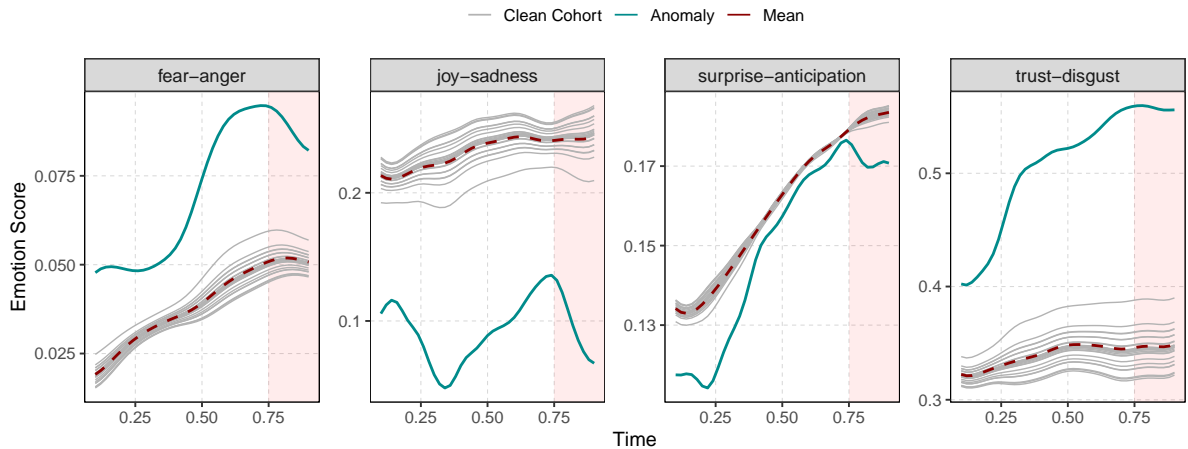

Figure 4: Mode of variation plot for a user along the fourth FPC (outlying) from cluster 3

stability.

Through mode-of-variation plots (see Section C.2 for details) and corresponding review excerpts in the flagged time window, we show that the detected anomalies capture customer pain points. Figure 4 shows a user's emotional trajectory relative to Cluster 3. The user's emotions are consistently shifted from the cluster mean along the fourth eigenfunction in Cluster 3, with a pronounced spike in the fear–anger petal and a sharp drop in joy–sadness during the final time window—signaling a clear pain point. Review texts from this period reveal issues with mismatched parts, specifically a replacement door-handle cover with incorrect keyhole cut-outs. The dominant complaints relate to product fit and quality control. These insights suggest actionable interventions, such as enforcing compatibility checks at purchase and improving final-stage quality control by the seller.

Figure 5 shows a user exhibiting a dip–recovery emotional pattern along the second eigenfunction. Early in the timeline, all four emotion petals remain well below the cluster baseline. During the anomalous time window, there is a sharp rise in fear and surprise, driven by issues related to poor product quality. The user expresses frustration and regret, suggesting loss of brand trust. Key pain points include the failure of a critical component and confusion caused by missing documentation.

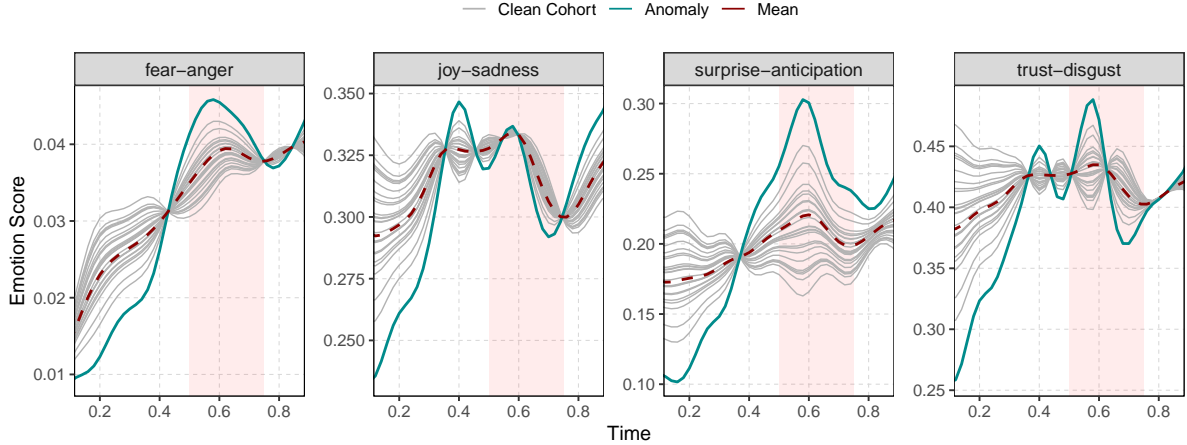

Figure 5: Mode of variation plot for a user along the second FPC (outlying) from cluster 1

**Keyword profiling** After detecting anomalies, we perform keyword profiling (Algorithm 11 in Section F of the Appendix) to each flagged instance. Table 2 summarizes the keywords associated with anomalous points in each cluster. A quick glance shows that users in Cluster 2 tend to express broadly negative emotions, while Cluster 3 highlights more specific issues—such as missing cables and poor documentation—reflecting the more descriptive and varied nature of reviews in that group. Table 6 illustrates dynamic profiling of keywords; see Section F in the Appendix for details.

| Cluster | Keywords |
| --- | --- |
| Cluster 1 | as described, good quality, perfect product, poor value for money, wrong size, poor fit |
| Cluster 2 | poor quality, poor value for money |
| Cluster 3 | as described, bulky design, good quality, good value for money, good design, leaks fuel, missing cable, quantity issue, poor documentation, wrong size |

Table 2: Group-level pain points detected across clusters

## 4.2 Tracking Toxicity and Aggression in Wikipedia request–comment stream

We evaluate LLmFPCA-detect on the English Wikipedia request–comment stream to demonstrate cross-domain applicability. For each comment, we record the text, timestamp, structured user covariates, and crowdsourced ground-truth toxicity/aggression scores. This corpus exemplifies sparse longitudinal text: users post at irregular, infrequent intervals. The dataset was collected via the Wikipedia API, restricted to the user–talk and article–talk namespaces, and sourced from Wiki data. We retain comments from 2010–2015 authored by 925 pseudonymized users.

| Method | TW1 | TW2 | TW3 | TW4 | TW5 |
| --- | --- | --- | --- | --- | --- |
| LLmFPCA-detect (gpt-4o-mini) | **0.58** | **0.58** | **0.46** | **0.37** | 0.32 |
| Isolation Forest (BERT) | 0.41 | 0.33 | 0.25 | 0.23 | 0.39 |
| Isolation Forest (gpt-4o-mini) | 0.41 | 0.33 | 0.25 | 0.23 | 0.39 |

Table 3: F1 scores for anomalies detected by LLmFPCA-DETECT versus ground truth, compared with Isolation Forest on GPT-derived scores and a BERT baseline (segregated by time windows).

**Comparison with state-of-the-art** We assess anomaly detection on Wikipedia by treating human-annotated toxicity/aggression as surrogate ground truth and extracting GPT-derived toxicity/aggression scores from text via prompts. As a content-agnostic baseline, we use BERT embeddings (no explicit toxicity cues). We partition the timeline into five windows and, within each, define pseudo–ground-truth anomalies using Isolation Forest on the human scores plus user covariates (comment count, median inter-comment gap). We then run Isolation Forest on (i) GPT-derived scores and (ii) BERT embeddings (each with the same covariates) as baselines. Finally, we apply LLmFPCA-detect to the GPT-derived trajectories with the same covariates to flag anomalies across the five windows and compare against these baselines (Table 3).

**Cluster dynamics** LLmFPCA-detect flags not only one-off vandalism or brief flare-ups by otherwise well-behaved contributors, but also sustained problematic behavior and its mode of deviation. For example, Cluster 1 outliers tend to post unusually high volumes or engage in extended policy disputes, whereas Cluster 2 outliers show short, intense bursts of toxic language. Table 4 presents representative cases with brief excerpts and the corresponding anomalous time window. In Cluster 1, the dominant pattern is procedural friction—disagreements about process (e.g., whether a proposed mentorship program requires further consensus) rather than direct attacks. By contrast, Cluster 2 features overt hostility, where procedural disagreements escalate into personal or con-frontational language. Additionally, Appendix D.3 reports bootstrap analyses confirming stability of the obtained clusters.

| Cluster | User ID | Comment excerpt (abridged) | Label |
|---|---|---|---|
| 1 | 10783082 | "…If that's how you want it. I will talk to this to ANI if necessary …" | 1 |
| 1 | 10756369 | "=== Adopt Me === Here is a proposal for a new mentorship process …" | 1 |
| 2 | 2305952 | "OK, maybe I was wrong. I'm sorry, but don't try me again …" | 5 |
| 2 | 2305952 | "No, that's irrelevant. Your source is garbage, stop spamming it." | 5 |

Table 4: Examples from the Wikipedia comment stream where detected anomalies match crowd-sourced annotations, showing cluster ID, anonymized user ID, excerpt, and toxicity/aggression label.

| Cluster 1 (Window) | Top keywords (LLmFPCA-detect) | Theme |
|---|---|---|
| W1 | consensus, policy, "WP: ANI" | Policy enforcement friction |
| W2 | civility, manners, please, courtesy | Soft-skills reminders |
| W3 | backlog, deadline, stall, formalise | Procedural urgency |
| **Cluster 2 (Window)** | **Top keywords (LLmFPCA-detect)** | **Theme** |
| W4 | nonsense, garbage-source, stop-spamming | Direct hostility |
| W5 | revert, vandal, warning, block, "3RR" | Conflict over content |
| W5 | wasting-time, already-explained | Moderator fatigue |

Table 5: Dynamic–keyword profiling makes each anomaly legible. In this Wikipedia setting, instead of an opaque outlier score, the moderator sees the top keywords that drove the statistical flag.

**Keyword profiling** Dynamic keyword profiling makes each anomaly interpretable (Table 5). Rather than an opaque outlier score, moderators see the top terms that triggered the flag, revealing the concerns underlying anomalous behavior. In this corpus, Cluster 1 anomalies are predominantly procedural—e.g., disputes over which venue (WP:ANI, etc.) should adjudicate. Cluster 2, by contrast, exhibits explicit antagonism: personal attacks, contempt for sources ("garbage-source"), and edit-war jargon. The exasperation lexicon ("wasting time," "already explained") further signals moderator fatigue—an operational risk that steady-state toxicity metrics would miss.

Identifying peak–hostility windows (e.g., Window 5) with LLmFPCA-detect enables proactive moderation, such as temporarily throttling edits. In Cluster 1, the dominant issue is procedural friction, suggesting policy fixes like clearer closure rules or targeted sanctions. Keyword profiling pinpoints specific, time-bounded situations where light-touch actions can prevent rule violations and burnout. Linking time windows to salient terms reveals root causes and supports proportionate, domain-specific responses instead of one-size-fits-all bans.

# 5 Conclusion

LLmFPCA-detect provides an end-to-end framework for sparse longitudinal (SL) text by integrating LLM-embeddings with functional data analysis. LLmFPCA-detect tackles key challenges in such datasets—including sparsity, irregularity, noise, and semantic complexity—by embedding text into meaningful numeric representations, followed by mFPCA which is used for user segmentation, anomaly detection, and dynamic intent profiling across large SL text datasets, a setting that remains largely unaddressed in the literature. Applied to Amazon customer reviews, LLmFPCA-detect successfully uncovers emotion dynamics and identifies critical pain points in the customer journey, offering valuable insights for consumer analytics. We demonstrate the utility of LLmFPCA-detect on English Wikipedia request–comment stream to detect toxic comments, where the detected anomalies align well with crowdsourced human annotations. The flexibility of LLmFPCA-detect makes it applicable to other domains such as healthcare, education, and social media where SL text data is routine. Future work includes establishing theoretical guarantees based on mFPCA estimates rather than fully observed trajectories, and extending LLmFPCA-detect to other supervised and unsupervised tasks on SL text datasets.

# Appendix

The appendix contains supplementary text, figures, and additional results that support the main paper.

# A  Temporal Dynamics of Customer Intents Across Clusters and the Structure of Technical Proofs

This section highlights the evolving patterns of customer intents across the four time windows for each cluster (Table 6) and directs readers to the Technical Appendices D.1 and E.1 for the formal proofs underpinning our clustering and anomaly-screening methods.

Table 6 presents the distribution of dominant customer intents—such as product quality, fit, and value perceptions—across the four sequential time windows for each of the three clusters. For each cluster, we list the most frequent intent labels along with their occurrence counts, highlighting how user concerns and satisfaction indicators evolve over the course of their interaction trajectory. This breakdown reveals distinct temporal patterns in user feedback: Cluster 1 users progressively emphasize product perfection in later windows, Cluster 2 maintains a consistently low volume of quality complaints, and Cluster 3 displays a broad diversity of intents, including both positive and negative evaluations, particularly in the final window.

Table 6: Summary of Intents and Counts by Cluster and Time Window

| Cluster | Time Window 1 | Time Window 2 | Time Window 3 | Time Window 4 |
| --- | --- | --- | --- | --- |
| Cluster 1 | poor value for money (4), wrong fit (1) | good product quality (2), perfect product (50), poor product quality (1), wrong fit (1) | as described (6), good product quality (10), perfect product (21), poor product quality (2), poor value for money (7), wrong fit (1) | good product quality (1), perfect product (31), poor value for money (2), wrong fit (1) |
| Cluster 2 | poor product quality (1) | poor product quality (3), poor value for money (1) | poor product quality (1) | poor product quality (2) |
| Cluster 3 | poor product design (3), poor product quality (3) | great value for money (1), poor product quality (2) | poor product design (1), poor product quality (1) | as described (3), good product quality (16), great bargain (3), great design (1), great value for money (10), missing charging cable (1), perfect product (1), poor product design (14), poor product quality (18), poor value for money (8), size variation issue (1), too big and bulky (2), wrong fit (4) |

**Technical Appendices** Appendix D.1 quantifies how well Lloyd's $k$-means recovers the true clusters when the data contain a small fraction of anomalous points. Theorem D.1 shows that, if the cluster means are sufficiently separated ($\Delta$) and the normalized signal-to-noise ratio $r_k$ exceeds a computable threshold, the mis-clustering rate of the clean data drops to $\exp(-\Delta^2/16\sigma^2)$ after at most $4 \log N$ iterations, starting from a modestly accurate initialization.

Appendix E.1 supplies the statistical basis for the screening and calibration steps used after clustering. It models clean mFPC scores as draws from a baseline law $\mathcal{L}_m^k$ and treats anomalies as score shifts by an independent random effect. Theorem E.1 proves that, when the within-cluster anomaly fraction $\pi_{a,k} = o(1)$, a simple tail test on the empirical score distribution always includes the anomalous subjects with high probability. These two results jointly justify the clustering–then–screening strategy used throughout the paper.

# B Experimental Setup, Simulations and Supplementary Results

## B.1 Computational Environments and Tasks

All emotion scoring and related language-model tasks were performed on setup 1: the Standard NC4as T4 v3 VM with a Tesla T4 GPU (16 GB), 4 vCPUs, 28 GB RAM, Linux (x64, Gen 1). All subsequent work—fitting algorithms, analyzing Amazon reviews, other experiments, simulations etc. —was carried out on setup 2: MacBook M3 Pro with 36 GB memory, and setup 3: 13 inch 2020 MacBook Pro (2.3 GHz Quad-Core i7, 32 GB LPDDR4X, macOS Sequoia 15.0.1). The full research project did not require more compute than the experiments reported in the paper.

## B.2 Simulations

For each cluster $\ell \in \{1, 2, 3\}$, we assess the stability of our anomaly-detection pipeline under realistic sparsity by performing a simulation study with $S = 50$ replicates on the Amazon Automobile review data (Section 4.1). Let $\hat{\mathcal{C}}_\ell$ be the set of users in cluster $\ell$ (from Algorithm 5), and let $\mathcal{A}^{(\ell)} = \{i \in \hat{\mathcal{C}}_\ell : A_i^{(\ell)} = 1\}$ denote the "base" anomalies originally flagged by Algorithm 2. We first obtain cluster-specific fitted emotion trajectories

$$\hat{X}_{i,\hat{\mathcal{C}}_\ell}^{(j)}(t), \quad i \in \hat{\mathcal{C}}_\ell, \; j = 1, \ldots, 4,$$

via the mFPCA procedure of Section 3 (Algorithms 1, 6). In replicate $s$, for each user $i$ we draw a truncated Poisson subsample size

$$K_i^{(s)} \sim \text{Pois}(10) \text{ truncated to } \{5, \ldots, 15\},$$

then uniformly select $K_i^{(s)}$ timepoints from the common grid $\{t_1, \ldots, t_T\}$ to form

$$Lt_i^{(s)} = \{t_{i1}^{(s)}, \ldots, t_{iK_i^{(s)}}^{(s)}\}, \quad Ly_i^{(j),(s)} = \{\hat{X}_{i,\hat{\mathcal{C}}_\ell}^{(j)}(t) : t \in Lt_i^{(s)}\}.$$

We apply univariate Sparse FPCA independently to each $(Lt_i^{(s)}, Ly_i^{(j),(s)})$ (Algorithm 4), stack the resulting scores into the matrix $H^{(\ell,s)}$, and then perform multivariate FPCA (Algorithm 1) to obtain mFPC scores $\hat{\rho}_{i,m}^{(\ell,s)}$. We then re-run the exact screening-calibration anomaly detection (Algorithm 2), yielding binary flags

$$A_i^{(\ell,s)} \in \{0,1\}, \quad i \in \hat{\mathcal{C}}_\ell.$$

where

$$A_i^{(\ell,s)} = \begin{cases} 1, & \text{user } i \text{ flagged as anomalous in replicate } s, \\ 0, & \text{not flagged in replicate } s. \end{cases}$$

By comparing $\{A_i^{(\ell,s)}\}_{s=1}^S$ against the base set $\mathcal{A}^{(\ell)}$, we compute per-user detection frequencies—i.e. the proportion of replicates in which originally flagged anomalies remain detected—thus quantifying the robustness of our pipeline to the sparse, irregular sampling patterns inherent in customer review trajectories.

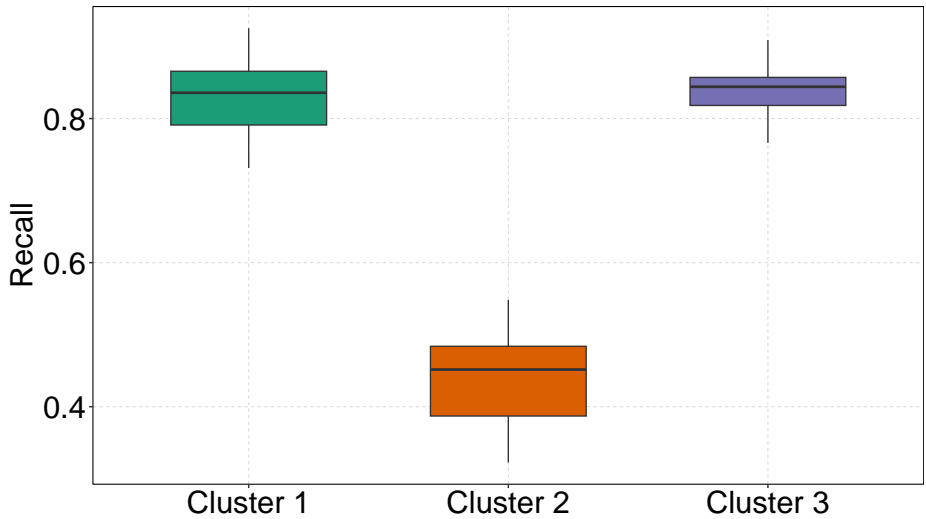

Figure 6: Distribution of recall across the clusters beginning from cluster 1 on the

Figure 6 plots, for each cluster $\ell$, the empirical distribution of

$$\text{Recall}_s^{(\ell)} = \frac{|\{i \in \mathcal{A}^{(\ell)} : A_i^{(\ell,s)} = 1\}|}{|\mathcal{A}^{(\ell)}|},$$

over $S = 50$ subsampling replicates $s$. Here

$$\mathcal{A}^{(\ell)} = \{i \in \hat{\mathcal{C}}_\ell : A_i^{(\ell)} = 1\}$$

is the set of "base" anomalies—i.e., the users flagged by Algorithm 2 on the full (unsparsified) trajectories—so $|\mathcal{A}^{(\ell)}|$ is the total number of those original anomalies in cluster $\ell$.

In Cluster 1 and Cluster 3 (the two larger clusters), the median recall exceeds 0.80 and the interquartile ranges are tight (appx. 0.78–0.92 and 0.82–0.94, respectively), indicating that over 80% of the base anomalies survive even when only 5–15 timepoints per user are observed. By contrast, Cluster 2 (the smallest cluster) has a markedly lower median recall (appx. 0.45) and a much wider spread (0.30–0.55), revealing that nearly half of its base anomalies are missed in many subsampled runs.

These differences reflect intrinsic heterogeneity in the temporal signatures of anomalous users. Clusters 1 and 3 appear to harbor anomalies whose deviations from the cluster mean are pronounced and sustained over time, so they survive aggressive subsampling. In Cluster 2, however, the outlying behavior is more localized or subtler, making detection highly sensitive to which windows are sampled. This suggests that, in practice, anomaly detection for Cluster 2 may benefit from (a) collecting additional observations around key time periods or (b) combining Type-1 score-based flags with complementary time-window tests (Algorithm 3) to recover those more fragile anomalies.

Figure 7 displays the per-user detection probability—i.e. the fraction of the 50 subsampling replicates in which each base anomaly is recovered—separately for the three clusters. In Cluster 1 (left panel), the density is strongly concentrated near 1.0, with a steep rise above 0.8, indicating that nearly all anomalies in this group are detected almost every time. Cluster 3 (right panel) exhibits a similar pattern, with a prominent mode around 0.9–1.0 and only a small tail below 0.7, confirming that its anomalies are likewise robust under sparse sampling. By contrast, Cluster 2's distribution (center panel) is centered around 0.3–0.4 and is much flatter, revealing that most of its anomalies are recovered in fewer than half of the replicates.

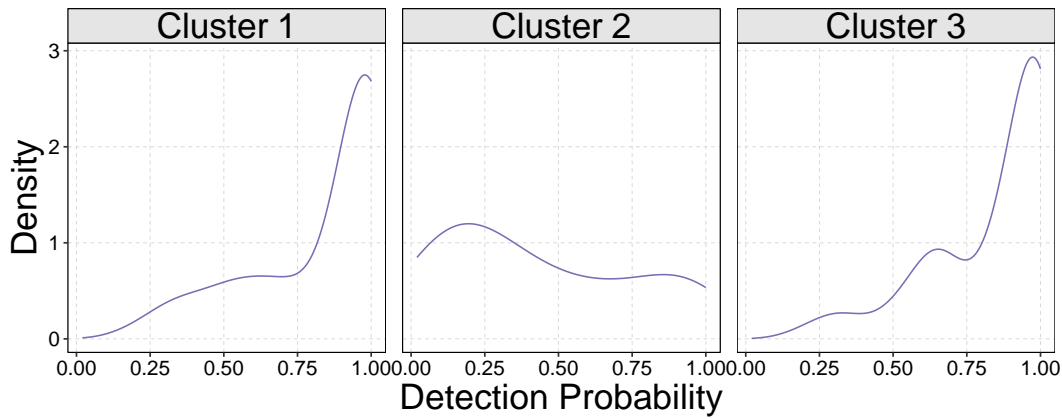

Figure 7: Detection probability across the clusters

These density profiles mirror our recall findings: anomalies in Clusters 1 and 3 produce large, sustained deviations from their cluster means and thus enjoy high, stable detection probabilities, whereas the subtler, more localized deviations in the smaller Cluster 2 lead to low and highly variable hit-rates. Together, these results underscore the need for enhanced detection strategies—such as targeted time-window testing (Algorithm 3)—to reliably capture the more fragile anomalies in Cluster 2.

**Recall Comparison Across Clusters: Significance Testing**  In order to formally evaluate whether the observed differences in recall across clusters are statistically significant, we employed a Kruskal–Wallis rank-sum test, a nonparametric test of significance for comparing more than two independent groups. Accounting for the imbalance in cluster sizes—Clusters 1 and 3 each contain approximately 800 users, whereas Cluster 2 contains only about 200—the null hypothesis $H_0$: "all three clusters have the same recall distribution" was tested at the $\alpha = 0.05$ level. We obtained

$$\chi^2(2) = 100.28, \quad p\text{-value} < 2 \times 10^{-16},$$

strongly rejecting $H_0$. To identify which clusters differ, we conducted pairwise Wilcoxon rank-sum tests with Benjamini–Hochberg adjustment for multiple comparisons. The adjusted p-values for

Cluster 2 vs. Cluster 1 and Cluster 2 vs. Cluster 3 were both

$$p_{\text{adj}} < 2 \times 10^{-16},$$

indicating highly significant lower recall in the small Cluster 2, while the comparison between Clusters 1 and 3 yielded

$$p_{\text{adj}} = 0.20,$$

consistent with no significant difference at the 5% level. Finally, to account for the paired nature of the 50 simulation runs, we applied a Friedman test, another nonparametric test of significance for related samples, which yielded

$$\chi^2(2) = 75.16, \quad p\text{-value} < 2 \times 10^{-16}.$$

Together, these statistical tests of significance confirm that Clusters 1 and 3 share equivalently high and stable recall under sparse sampling, whereas the much smaller Cluster 2 exhibits a significantly and substantially weaker recall performance.

**Supplemental details regarding experiments perform in Section 4.1:** For Cluster 2 we applied the two-stage anomaly pipeline with screening threshold $\alpha_1 = 0.15$ and Bonferroni-corrected confirmation $\alpha = 0.10$ across $B = 4$ mFPC loadings, followed by window-wise testing at $\alpha = 0.20$ across $W = 4$ fixed intervals. All confirmed outliers $i \in \mathcal{A}^{(2)}$ were then summarized in the Excel file `cluster2_pvalues.xlsx`.

This spreadsheet has nine columns:

- `user_id`: the unique subject identifier $i$.

- `p_comp1`, ..., `p_comp4`: the Bonferroni-adjusted empirical p-values $\hat{p}_{i,m}^{\text{emp}}$ from the Type-1 test on the $m$th mFPC loading ($m = 1, \ldots, 4$).

- `p_win1`, ..., `p_win4`: the Bonferroni-adjusted empirical p-values $\hat{p}_i^{(w)}$ from the Type-2 test in the $w$th time window ($w = 1, \ldots, 4$).

Entries marked `NA` in the `p_win*` columns indicate windows with no observations for subject $i$, while numeric values record the corresponding adjusted p-value. This table therefore provides, for each of the 56 Cluster 2 anomalies, a complete vector of significance measures across both the principal-component and time-window analyses.

For Clusters 1 and 3, we applied the two-stage anomaly pipeline with screening threshold $\alpha_1 = 0.1$ and Bonferroni-corrected confirmation $\alpha = 0.05$ across $B = 4$ mFPC loadings, followed by window-wise testing at $\alpha = 0.10$ across $W = 4$ fixed intervals. Similar details are available for clusters 1 and 3 in files `cluster1_pvalues.xlsx` and `cluster3_pvalues.xlsx` respectively.

The Amazon reviews data used for both experiments and simulations is sourced from Amazon customer reviews datasets and selecting the appropriate item categories as mentioned in the paper.

Prompts for emotion-scoring implementation using Plutchik's wheel is provided in Section C.

Further methodological specifics—including clustering and anomaly-detection algorithms, theoretical underpinnings, subroutine descriptions, keyword profiling procedures, and emotion-scoring validation—are detailed in Appendix Sections C through G.

# C Supplemental details for Emotion scoring and validation and FPCA Methodology

Our primary analysis centers on the Automobile category of the Amazon reviews dataset, which includes 1,946 users and 22,032 reviews over five years. Reviews from the other two categories are used to construct user-level covariates, such as the proportion of purchases across categories. Each review includes a user ID, timestamp, product title, review text, and a 1–5 star rating. Users post reviews at irregular intervals over multiple years.

## C.1 Emotion scoring and validation based on Plutchik's wheel of emotions

The extraction of emotions from customer reviews and interaction transcripts is critical for organizations seeking to move beyond coarse sentiment analysis and gain actionable insights into the nuanced affective states of their users. Idnetification and quantification of emotions embedded in textual feedback allows businesses to tailor their responses, segment customers, and proactively address emerging issues. Among various emotion detection frameworks, Plutchik's Wheel of Emotions is frequently chosen due to its structured and psychoevolutionary foundation, which captures both the complexity and gradation of human affect Plutchik [1980], Seconds [2025], Semeraro et al. [2021]. In natural language processing, researchers have utilized Plutchik's model to relabel emotion datasets, enabling more nuanced detection and classification of emotions in text, which leads to improved performance in emotion recognition tasks, especially for complex or subtle affective states Mohammad and Bravo-Marquez [2018], Alhuzali and Ananiadou [2021]. Beyond text, Plutchik's model also informs emotion AI applications in facial recognition and user experience design, helping systems identify, interpret, and appropriately respond to the emotional states of users, thereby enhancing personalization and emotional intelligence in human-computer interaction Calefato et al. [2017]. Unlike continuous dimensional models or basic discrete emotion sets, Plutchik's wheel organizes eight core emotions—joy, trust, fear, surprise, sadness, anticipation, anger, and disgust—along with their intensities and combinations, enabling a richer and more interpretable taxonomy for emotion detection Seconds [2025], Semeraro et al. [2021]. This is represented in 4 petals with opposing extremes: joy-sadness, trust-fear, surprise-anticipation, anger-disgust. The wheel's compositional structure also facilitates algorithmic implementation, allowing AI systems to process negations and blends of emotions through arithmetic and logical operations, which is particularly advantageous for analyzing customer feedback, social media, and conversational data Mohammad [2020]. Leveraging Plutchik's model thus ensures a comprehensive and interpretable framework for mapping textual cues to specific emotional categories and intensities, facilitating more robust and actionable emotion analytics than alternative approaches.

In this work, we use in-context learning with to prompt GPT-3.5-Turbo to map the customer reviews to each of the 4 dimensions corresponding to Plutchik's Wheel of Emotions

Listing 1 and table 7 outline the scores and the prompt used for "joy–sadness" petal. Emotions are mapped to a continuous scale from −1 to 1, where positive values represent varying intensities of joy (serenity, joy, ecstasy) and negative values represent varying intensities of sadness (pensiveness, sadness, grief). Similar prompts and scores are used for other petals as well. The rubric is as follows:

| Emotion | Intensity | Score Range |
| --- | --- | --- |
| Ecstasy | Intense | $(\frac{2}{3}, 1]$ |
| Joy | Moderate | $(\frac{1}{3}, \frac{2}{3}]$ |
| Serenity | Mild | $(0, \frac{1}{3}]$ |
| Pensiveness | Mild | $[0, -\frac{1}{3})$ |
| Sadness | Moderate | $[-\frac{1}{3}, -\frac{2}{3})$ |
| Grief | Intense | $[-1, -\frac{2}{3})$ |

Table 7: Emotion scoring rubric based on Plutchik's joy–sadness petal.

The following prompt is used to guide the scoring process for each review:

**Listing 1:** *Prompt for Plutchik-based Emotion Scoring*

```
You are an expert at honestly classifying and scoring emotion from text.
This task involves analyzing user's review of a product, consisting of a review_title
    and review_text.
The objective is to quantitatively measure emotions, and score them based on a
    structured framework of emotions on a scale of -1 and 1.

Emotions can be represented using Plutchik's wheel of emotion using 4 pairs of petals
    and opposing petal combinations.
For this task, focus on one petal and its opposing pair. The petal includes: Ecstacy,
    joy and serenity.
The opposing petal includes: Grief, sadness and pensiveness. The emotions on each petal
    are arranged on the basis of their intensities.
Ecstacy is the most intense form of emotion and its opposing pair is grief. Joy is a
    moderate form and its opposing pair is sadness.
Similarly Serenity is a mild form, and pensiveness is its opposing pair.

--- INSTRUCTIONS ---
Carefully read the user's review, including both the review_title and review_text. Based
```

```
     on the emotional cues present,
use the structured framework of the emotions, described below, along with their
    respective intensities, to guide your analysis,
and score the review, on a scale of -1 and 1.

The emotion score should lie between 0 and 1 if the review suggests ecstacy, joy or
    serenity, while it should lie between 0 and -1
if the review suggests grief, sadness or pensiveness. The score should be determined
    based on the intensity of the emotion. For e.g.
since ecastacy is the most intense form of emotion it should be scored between 2/3 and
    1.
Similarly Grief being the most intense opposing pair, it should be scored on a scale of
    -2/3 and -1. Similarly if joy is detected it
should be scored on a scale of 1/3 and 2/3, and if its opposing pair sadness is detected
     it should be scored on a scale of -1/3 and -2/3.
Similarly if Serenity is detected it should be scored on a scale of 0 and 1/3 and if its
     opposing pair pensiveness is detected it should
be scored on a scale of 0 and 1/3. Absence of these emotions defaults to a score of 0.
The final score should be a single number between -1 and 1, reflecting the emotion and
    its intensity.

Petal Dynamics:
   Joy: Indicates happiness or pleasure derived from product satisfaction.
     - Serenity (Mild)
     - Joy (Moderate)
     - Ecstasy (Intense)

Opposing Petal Dynamics:
   Sadness: Reveals disappointment or sorrow due to unmet product expectations.
     - Pensiveness (Mild)
```

```
    - Sadness (Moderate)
    - Grief (Intense)

--- TASK ---
For each user's review_title and review_text, follow the instructions to score the
    emotion expressed
on a scale of -1 and 1. Ensure the final score is a single number between -1 and 1 based
    on petal or opposing petal dynamics, without any additional explanation.
in the format: 'Score= '
```

**Scoring Emotions of Amazon reviews based on Plutchik's Wheel of Emotions, and Mode of Variation plots** Throughout the empirical study we specialise to $p = 4$ by letting $\Phi$ return continuous scores along the four opposing-petal pairs of Plutchik's Wheel of Emotions. Concretely,

$$Y_i(T_{ij}) = \left(e_{ij}^{(1)}, e_{ij}^{(2)}, e_{ij}^{(3)}, e_{ij}^{(4)}\right), \tag{C.1}$$

where, for instance, $e_{ij}^{(1)} = +1$ encodes intense *Joy* and $e_{ij}^{(1)} = -1$ encodes intense *Sadness*, with proportionate grading in between. Appendix G reports the prompt template and a manually–annotated validation set showing an average match rate of approximately 56 %.

Let each customer be indexed by $i \in \{1, \ldots, N\}$, and let $x_{ij}$ denote the textual datapoint (e.g., a review, comment, or message) associated with the $j$-th interaction of customer $i$, recorded at timestamp $T_{ij} \in \mathcal{T}_i \subset \mathbb{R}$, where $j \in \{1, \ldots, m_i\}$ and $m_i$ is the number of observed interactions for customer $i$. Each $x_{ij}$ is mapped to a four-dimensional vector of continuous emotion scores:

$$\mathbf{e}_{ij} = \left(e_{ij}^{(1)}, e_{ij}^{(2)}, e_{ij}^{(3)}, e_{ij}^{(4)}\right) \in [-1, 1]^4, \tag{C.2}$$

where each component $e_{ij}^{(d)}$ for $d \in \{1, 2, 3, 4\}$ corresponds to an opposing-petal pair in Plutchik's Wheel of Emotions: Joy–Sadness, Anger–Fear, Trust–Disgust, and Anticipation–Surprise, respectively.

The value $e_{ij}^{(d)} = +1$ indicates strong expression of the first emotion in the pair (e.g., Joy), $-1$

denotes strong expression of the second (e.g., Sadness), and 0 corresponds to emotional neutrality.

Emotion scores are generated by querying GPT-3.5 Turbo leveraging prompt engineering with a fixed prompt that defines each Plutchik pair and instructs the model to output four scalar intensities representing the emotional content of $x_{ij}$ on the $[-1, 1]$ scale. This pipeline produces a deterministic mapping from each textual interaction to an interpretable, low-dimensional emotional representation. Although each dimension is extracted independently, and therefore modeled as marginally uncorrelated at this stage, we do not assume that the underlying emotional processes are statistically independent. In particular, we later capture their joint evolution using multivariate sparse functional models (Section 3).

**Validation** To validate our numeric embedding scheme, we randomly sampled 30 reviews from across categories and obtained expert human labels for the 24 emotion categories across the four Plutchik petal and opposing petal combinations. We then ran a GPT prompt 2 on those reviews, performed classification, and computed exact-match accuracy ($19/33 = 57.6$ %). Accuracy was measured as the proportion of reviews for which GPT's emotion classification matched the consensus human annotation (see Appendix G, Table 8).

Formally, for each customer $i$ and emotion dimension $d$, we define the observed emotion signal as a sparse, irregular time series:

$$\mathcal{E}_i^{(d)} = \left\{ \left(T_{ij}, e_{ij}^{(d)}\right) : j = 1, \ldots, m_i \right\}, \quad d = 1, \ldots, 4. \tag{C.3}$$

The full sequence $\{\mathbf{e}_{ij}\}_{j=1}^{m_i}$ defines a multivariate, irregularly sampled emotional trajectory for customer $i$, which serves as the input to the functional data analysis pipeline developed in subsequent sections.

Evaluation of the emotion scoring method, including prompt design and accuracy computed against manually annotated examples, is described in Appendix G.

**Mode of variation plots** This routine produces a series of faceted line plots that contrast each confirmed anomalous subject's component-driven trajectory deviations with both the cluster mean and a reference cohort along a given anomalous eigenfunction direction $\boldsymbol{\psi}_m$.

For a subject $i \in \mathcal{C}$ flagged by Algorithm 2 and each anomalous direction $m \in S_i$, the following curves are drawn for each embedding dimension $d = 1, \ldots, p$:

1. Cluster Mean: $\hat{\mu}_{\hat{\mathcal{C}}}^{(d)}(t)$, as estimated in Algorithm 6.

2. Clean Cohort trajectories: $\hat{\mu}_{\hat{\mathcal{C}}}^{(d)}(t) + \hat{\rho}_{jm,\hat{\mathcal{C}}} \hat{\psi}_{m,\hat{\mathcal{C}}}^{(d)}(t)$ for each reference subject $j \in G_s^c$ (the clean held-out set from Algorithm 7).

3. Subject trajectory: $\hat{\mu}_{\hat{\mathcal{C}}}^{(d)}(t) + \hat{\rho}_{im,\hat{\mathcal{C}}} \hat{\psi}_{m,\hat{\mathcal{C}}}^{(d)}(t)$.

Only interior time-points are displayed to avoid boundary artifacts. If window-level anomalies $\mathcal{W}_i \subset \{1, \ldots, W\}$ were identified by Algorithm 10, the corresponding intervals on the time-axis are lightly shaded. Line styling distinguishes the cohort (thin solid), the anomalous subject (thicker solid), and the mean (dashed). The outcome is a list of $B$ plots—one per anomalous component—each facetted by dimension $d$, offering a clear visual summary of how the subject's trajectory departs from typical variation.

**Cluster 2 examples** On top of the existing results in Section 4.1 we provide additional examples leveraging mode of variation plots for users in cluster 2. Users in cluster 2 generally have predominantly lower scores on the joy and trust petal as evident from figure 3. Figure 8 represents a user who is no different, and has even more extreme emotion trajectory resulting in being flagged as an outlier. The trajectory is decisively negative—missing parts, poor cleaning performance, and cheaply made accessories drive Fear–Anger and Surprise–Anticipation to their highest levels while Trust–Disgust turns consistently downward. The final window of the journey remains dominated by pain points that describe products falling off, tearing during installation, or arriving fused or broken; correspondingly, all four emotion curves sit well below the cluster mean, and the user is flagged as an outlier for extreme negativity. Potential actionable insights include tightening vendor quality control and shipping inspection for fragile SKUs, adding explicit fit-compatibility checks at checkout, and prioritizing fast replacement for early-life failures—measures that directly target the defects driving the user's strongest negative reactions.

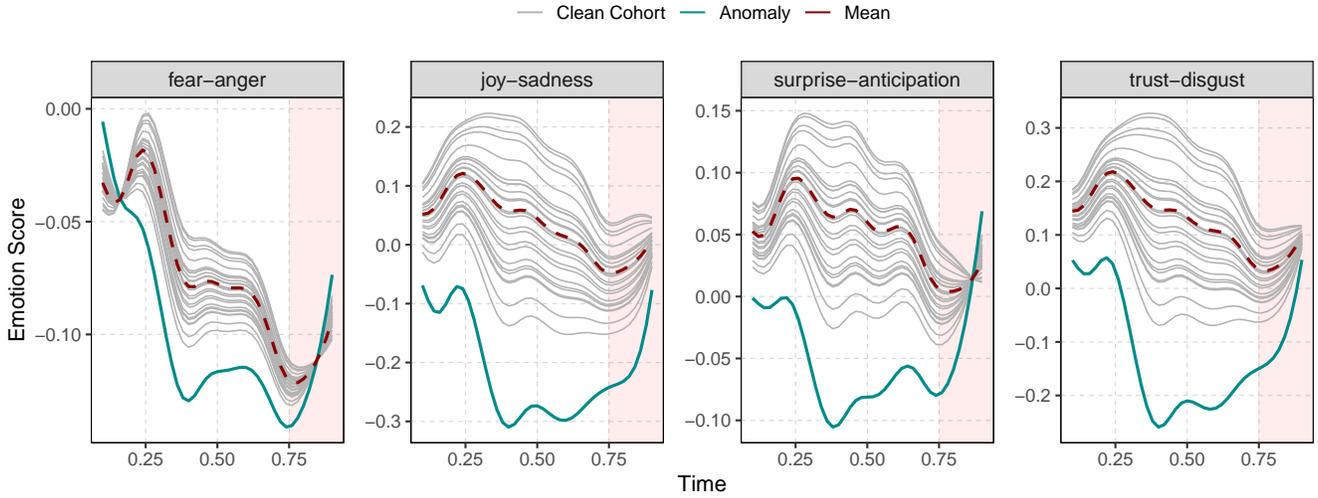

Figure 8: Mode of variation plot for a user along the first FPC (outlying) from cluster 2

Given that the cluster has predominantly negative emotions, one would expect a user within this cohort experiencing positive product experiences at some point in their journey to be an outlier. This is exactly what we observe for the user depicted by the emotion customer journey curves, as in the figure 9. This further showcases the validity of our anomaly detection algorithm where the outlying customers are detected with respect to cluster means. It can be observed that the joy petal increases significantly towards the later half of the customer's journey. In the first half of the timeline, the user echoes the cluster mood: adhesive patches fail, and an overpriced organiser proves hard to fold. Fear–Anger and Surprise–Anticipation rise sharply here, while Joy–Sadness and Trust–Disgust sink, mirroring these early disappointments. Beginning at roughly the midpoint, the pattern reverses: successive five-star reviews report magnetic trays that simplify sewing, door-edge guards installed from a wheelchair without scratches, and a series of easy-to-use safety and cleaning accessories. All positive emotion curves swing upward, marking the user as a positive outlier against an otherwise pessimistic cohort.

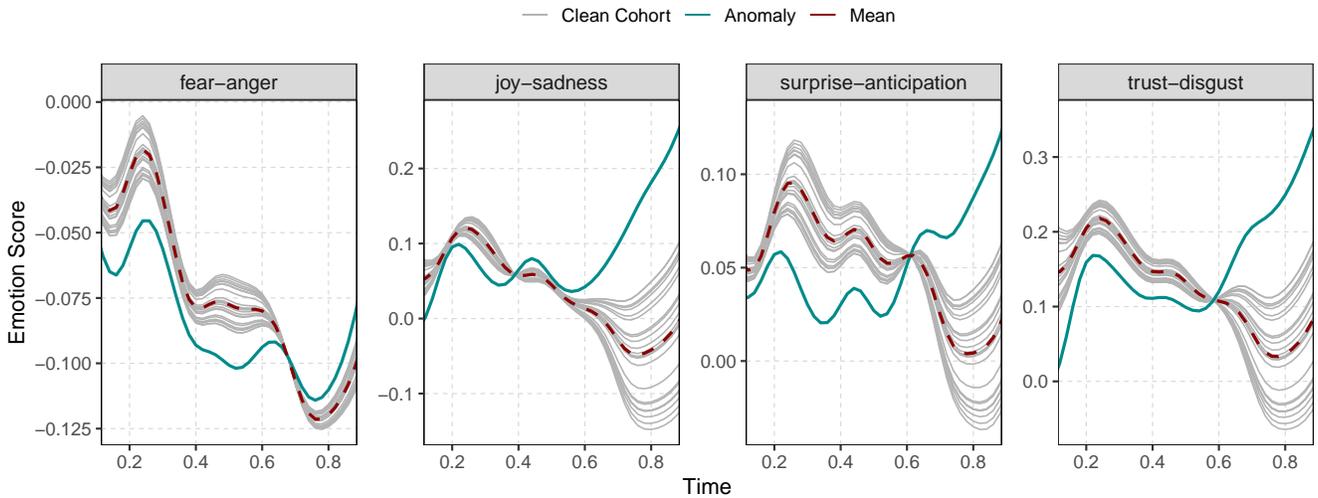

Figure 9: Mode of variation plot for a user along the second FPC (outlying) from cluster 2

## C.2  Methodological framework for univariate and multivariate FPCA

Univariate FPCA models each embedding dimension separately, yet the coordinates of a subject's numeric trajectory often exhibit strong dependencies. To capture their joint variation, we employ Multivariate Sparse Functional PCA following the framework of Happ et al. (2018). Let $\hat{\xi}_{ik}^{(d)}$ denote the demeaned univariate FPC scores for subject $i$ on the $k$th basis of dimension $d = 1, \ldots, p$. We assemble these into the score matrix

$$\Xi \in \mathbb{R}^{N \times M}, \quad \text{with rows } \Xi_i = \big(\hat{\xi}_{i1}^{(1)}, \ldots, \hat{\xi}_{iK_1}^{(1)}, \ldots, \hat{\xi}_{i1}^{(p)}, \ldots, \hat{\xi}_{iK_p}^{(p)}\big),$$

where $M = \sum_{d=1}^{p} K_d$. The sample covariance of the stacked scores is

$$\widehat{Z} = \frac{1}{N-1} \Xi^\top \Xi \in \mathbb{R}^{M \times M},$$

and its eigen-decomposition yields eigenpairs $(\hat{\nu}_m, \hat{c}_m)$ for $m = 1, \ldots, M$. The resulting multivariate FPC scores for subject $i$ are

$$\hat{\rho}_{im} = \Xi_i \cdot \hat{c}_m, \qquad i = 1, \ldots, N.$$

To reconstruct smooth, low-dimensional trajectories, we project each univariate eigenfunction $\hat{\phi}_n^{(d)}(t)$ onto the multivariate basis:

$$\hat{\psi}_m^{(d)}(t) = \sum_{n=1}^{K_d} \hat{c}_{m,n}^{(d)} \, \hat{\phi}_n^{(d)}(t),$$

where $\hat{c}_{m,n}^{(d)}$ denotes the entry of $\hat{c}_m$ corresponding to dimension $d$. The centered reconstruction on dimension $d$ is then

$$\widetilde{X}_i^{(d)}(t) = \sum_{m=1}^{M} \hat{\rho}_{im}\, \hat{\psi}_m^{(d)}(t),$$

and adding back the univariate mean $\hat{\mu}^{(d)}(t)$ yields the final multivariate embedding

$$\widehat{X}_i^{(d)}(t) = \hat{\mu}^{(d)}(t) + \sum_{m=1}^{M} \hat{\rho}_{im}\, \hat{\psi}_m^{(d)}(t), \quad d = 1, \ldots, p.$$

This procedure delivers a smooth, joint summary of each subject's $p$-variate trajectory while preserving the original scale of the numeric embeddings.

After extracting multivariate FPC scores $\hat{\rho}_{im}$ for $m = 1, \ldots, M$, we form the score vector $\boldsymbol{\rho}_i = (\hat{\rho}_{i1}, \ldots, \hat{\rho}_{iM})$ for each subject $i$. Time-invariant covariates (e.g. average rating, review length) are collected in $\mathbf{z}_i \in \mathbb{R}^q$. We concatenate these into joint features

$$\mathbf{w}_i = (\boldsymbol{\rho}_i, \mathbf{z}_i) \in \mathbb{R}^{M+q},$$

then center and whiten them by computing

$$\bar{\mathbf{w}} = \frac{1}{N}\sum_{i=1}^{N} \mathbf{w}_i, \quad W_c = \begin{bmatrix} \mathbf{w}_1 - \bar{\mathbf{w}} \\ \vdots \\ \mathbf{w}_N - \bar{\mathbf{w}} \end{bmatrix}, \quad \Sigma = \frac{1}{N-1} W_c^\top W_c, \quad \tilde{\mathbf{w}}_i = \Sigma^{-1/2}(\mathbf{w}_i - \bar{\mathbf{w}}),$$

so that $\{\tilde{\mathbf{w}}_i\}$ have zero mean and identity covariance. Clustering is then performed on $\{\tilde{\mathbf{w}}_i\}$ yields groups $\{\mathcal{C}_\ell\}_{\ell=1}^{L}$ of subjects with similar joint profiles using algorithm 5 in section 6 below.

**Univariate Functional Principal Component Analysis (uFPCA) Subroutine** From Section C.2, we have that we model each subject's latent, continuous embedding trajectory in dimension $d = 1, \ldots, p$ by

$$X_i^{(d)}(t), \quad t \in \mathcal{T}, i = 1, \ldots, N.$$

Each observed embedding

$$Y_i^{(d)}(T_{ij}) = X_i^{(d)}(T_{ij}) + \varepsilon_{ij}^{(d)}, \qquad \varepsilon_{ij}^{(d)} \sim \mathcal{N}(0, \sigma_d^2),$$

thus forms sparse functional data due to the irregular and limited sampling times $\{T_{ij}\}_{j=1}^{N_i}$.

To recover the main modes of variation in each dimension separately, we apply the univariate sparse FPCA of Yao et al. [2005] using algorithm 4 below.

Subroutine algorithm 4 (uFPCA): This subroutine implements the univariate sparse functional principal component analysis on irregularly sampled observations. Starting from the set of pairs $\{(T_{ij}, Y_{ij})\}$, it first estimates the mean trajectory $\hat{\mu}(t)$ (for example, via local polynomial smoothing) and then constructs a smoothed covariance surface $\hat{G}(s,t)$. An eigen-decomposition of $\hat{G}$ yields the principal component functions $\{\hat{\phi}_k(t)\}$ and their associated variances $\{\hat{\lambda}_k\}$. The number of components $K$ is chosen—typically by the fraction of variance explained—and the subject-specific FPC scores $\{\hat{\xi}_{ik}\}$ are estimated (for example, using the PACE algorithm), with centering to ensure mean zero. The outputs comprise the centered scores, the estimated mean function, and the eigenfunctions.

---

**Algorithm 4** `uFPCA`: Univariate Functional Principal Component Analysis (uFPCA) Subroutine.

**Input:** Univariate sparse functional data $\{(T_{ij}, Y_{ij}) : i = 1, \ldots, N, j = 1, \ldots, N_i\}$.

1: Estimate mean function $\hat{\mu}(t)$. ▷ e.g., via local polynomial smoothing
2: Estimate covariance surface $\hat{G}(s,t)$. ▷ e.g., via smoothing of raw covariances
3: Perform eigen-decomposition of $\hat{G}(s,t)$ to get eigenfunctions $\hat{\phi}_k(t)$ and eigenvalues $\hat{\lambda}_k$.
4: Determine number of components $K$ (if not pre-specified). ▷ e.g., using Fraction of Variance Explained (FVE)
5: Estimate FPC scores $\hat{\xi}_{ik}$ for $k = 1, \ldots, K$, ensuring they are centered. ▷ e.g., via PACE Yao et al. [2005]

**Output:** Tuple: (Centered FPC scores $\{\hat{\xi}_{ik}\}_{i=1,k=1}^{N,K}$, mean function $\hat{\mu}(t)$, eigenfunctions $\{\hat{\phi}_k(t)\}_{k=1}^{K}$).

---

For $d = 1, \ldots, p$, we approximate

$$X_i^{(d)}(t) \approx \mu^{(d)}(t) + \sum_{k=1}^{K_d} \xi_{ik}^{(d)} \phi_k^{(d)}(t),$$

where $\mu^{(d)}(t)$ is the population mean function, $\{\phi_k^{(d)}(t)\}$ are the principal component functions, and $\{\xi_{ik}^{(d)}\}$ are the corresponding FPC scores for subject $i$. The truncation level $K_d$ is chosen by the proportion of variance explained.

Denoting the estimates by $\hat{\mu}^{(d)}$, $\hat{\phi}_k^{(d)}$, and $\hat{\xi}_{ik}^{(d)}$, the fitted trajectory for subject $i$ in dimension $d$ becomes

$$\hat{X}_i^{(d)}(t) = \hat{\mu}^{(d)}(t) + \sum_{k=1}^{K_d} \hat{\xi}_{ik}^{(d)} \, \hat{\phi}_k^{(d)}(t).$$

This provides a smooth, low-dimensional summary of each subject's $p$-variate embedding trajectory in its original scale.

# D Fundamentals of Clustering and the Corresponding Algorithmic Framework

## D.1 Recovery of Clusters

To analyze cluster recovery, we condition on $\boldsymbol{a}_i$ and treat it as deterministic. Assume the trajectories $\boldsymbol{Y}_i$ are fully observed for all $i = 1, \ldots, N$. Then, for non-anomalous subjects $i \in \mathcal{C}_k \cap \mathcal{A}_0^C$, , we have $\mathbb{E}(\boldsymbol{Y}_i(t)) = \boldsymbol{\mu}_k(t)$ for all $t \in \mathcal{T}$, whereas for anomalous subjects $i \in \mathcal{C}_k \cap \mathcal{A}_0$, $\mathbb{E}(\boldsymbol{Y}_i(t)) = \boldsymbol{\mu}_k(t) + \boldsymbol{a}_i(t)$ for all $t \in \mathcal{T}$. Define the overall population mean of $\boldsymbol{X}_i$ as $\boldsymbol{\mu}^\star(t) = \sum_{k=1}^K \pi_k \boldsymbol{\mu}_k(t)$, for all $t \in \mathcal{T}$ with $\pi_k$ being the cluster proportions. The corresponding population mean of $\boldsymbol{Y}_i$ is then given by $\boldsymbol{\mu}(t) = \boldsymbol{\mu}^\star(t) + \bar{\boldsymbol{a}}(t)$ where $\bar{\boldsymbol{a}}(t) = \frac{1}{N} \sum_{i=1}^N \boldsymbol{a}_i(t)$ for all $t \in \mathcal{T}$. Since the eigenfunctions $\boldsymbol{\psi}_m \in L^2(\mathcal{T})^p$ are shared across clusters, performing mFPCA on $\boldsymbol{X}_i - \boldsymbol{\mu}^\star$ using the pooled sample yields the same eigenfunctions $\boldsymbol{\psi}_m$, $m = 1, \ldots, M$, as the cluster-specific mFPCA. This is equivalent to performing mFPCA on $\boldsymbol{Y}_i - \boldsymbol{\mu}$, after adjusting the covariance operator by removing the diagonal noise component $\sigma_\eta^2$; see [Yao et al., 2005] for details. The mFPC scores of $\rho_{im}$ of $\boldsymbol{Y}_i - \boldsymbol{\mu}$ are then given by $\rho_{im} = \langle \boldsymbol{Y}_i - \boldsymbol{\mu}, \boldsymbol{\psi}_m \rangle$. One can decompose the mFPC scores as

$$\rho_{im} = \langle \boldsymbol{Y}_i - \boldsymbol{\mu}, \boldsymbol{\psi}_m \rangle = \langle \boldsymbol{X}_i - \boldsymbol{\mu}_k, \boldsymbol{\psi}_m \rangle + \langle \boldsymbol{\mu}_k - \boldsymbol{\mu}^\star, \boldsymbol{\psi}_m \rangle + \langle \boldsymbol{\eta}_i, \boldsymbol{\psi}_m \rangle + \langle \boldsymbol{a}_i - \bar{\boldsymbol{a}}, \boldsymbol{\psi}_m \rangle.$$

For $i \in \mathcal{C}_k \cap \mathcal{A}_0^C$, $\mathbb{E}(\rho_{im}) = \langle \boldsymbol{\mu}_k - \boldsymbol{\mu}^\star, \boldsymbol{\psi}_m \rangle$ as for $i \in \mathcal{C}_k \cap \mathcal{A}_0^C$, $\boldsymbol{a}_i \equiv \boldsymbol{0}$, and $\mathbb{E}(\langle \boldsymbol{X}_i - \boldsymbol{\mu}_k, \boldsymbol{\psi}_m \rangle) = 0$ and $\mathbb{E}(\langle \boldsymbol{\eta}_i, \boldsymbol{\psi}_m \rangle) = 0$ for all $i \in \mathcal{C}_k$. Similarly, for all $i \in \mathcal{C}_k \cap \mathcal{A}_0$, $\mathbb{E}(\rho_{im}) = \langle \boldsymbol{\mu}_k - \boldsymbol{\mu}^\star, \boldsymbol{\psi}_m \rangle + \langle \boldsymbol{a}_i - \bar{\boldsymbol{a}}, \boldsymbol{\psi}_m \rangle$. Thus, mFPCA applied to all trajectories $\boldsymbol{Y}_i$ produces scores whose expectations differ across clusters and reflect anomalies when present. This motivates modeling each subject's scores jointly with their static covariates $\boldsymbol{Z}_i$ via $\boldsymbol{w}_i = (\rho_{i1}, \ldots, \rho_{iM}, Z_1, \ldots, Z_q) \in \mathbb{R}^{M+q}$ as

$$\boldsymbol{w}_i = \boldsymbol{\theta}_k + \boldsymbol{\epsilon}_i + \boldsymbol{\alpha}_i, \quad \text{for } i \in \mathcal{C}_k,$$

where $\boldsymbol{\theta}_k \in \mathbb{R}^{M+q}$ is the cluster mean of the uncorrupted points in $\mathcal{C}_k$, $\boldsymbol{\epsilon}_i \in \mathbb{R}^{M+q}$ is a sub-Gaussian random vector with covariance $\Sigma \in \mathbb{R}^{M+q \times M+q}$ such that for $\sigma > 0$, $\mathbb{E}[\boldsymbol{\epsilon}_i] = \boldsymbol{0}$, and for all $\boldsymbol{v} \in \mathbb{R}^{M+q}$, $\mathbb{E}\left[\exp\left(\langle \boldsymbol{v}, \boldsymbol{\epsilon}_i \rangle\right)\right] \leq \exp\left(\frac{\sigma^2 \|v\|^2}{2}\right)$. The term $\boldsymbol{\alpha}_i$ represents perturbations due to anomalies, with $\boldsymbol{\alpha}_i = \boldsymbol{0}$ for $i \in \mathcal{A}_0^C$ and $\boldsymbol{\alpha}_i \neq \boldsymbol{0}$ otherwise. Since we are interested in anomalies in the functional trajectories, we assume that $\boldsymbol{\alpha}_i \neq \boldsymbol{0}$ only in the first $M$ coordinates for $i \in \mathcal{A}_0$, i.e. $\alpha_{ij} = 0$ for all $j \in \{M+1, \ldots, M+q\}$. Let $\mathbb{E}(Z_i) = \gamma_k$ for $i \in \mathcal{C}_k$.

Let $\Delta = \min_{k \neq \ell} \|\boldsymbol{\theta}_k - \boldsymbol{\theta}_\ell\|$ denote the signal strength (separation between the means of the uncorrupted data across clusters). In terms of FPC scores, $\theta_{km} = \langle \boldsymbol{\mu}_k - \boldsymbol{\mu}^\star, \boldsymbol{\psi}_m \rangle$ for $m = 1, \ldots, M$, and $\theta_{km} = \gamma_k$ for $m \geq M+1$. Let $N_k = |\mathcal{C}_k|$, and let $N_k^a = |\mathcal{C}_k \cap \mathcal{A}_0|$ denote the number of anomalies in cluster $\mathcal{C}_k$. Write $N_a = |\mathcal{A}_0|$, $N^\star = N - N_a$, and $\varepsilon = N_a/N$. We run trimmed $k$-means with trimming rate $\tau$ and then apply Lloyd's algorithm to the retained observations to estimate cluster assignments. The following argument makes the retained set essentially free of anomalies before Lloyd.

Let $d = M + q$. Fix $\delta \in (0, 1/2)$ and define the inlier $(1-\delta)$-radius $R_{1-\delta}$ by

$$\Pr\left(\|\boldsymbol{w}_i - \boldsymbol{\theta}_k\| \leq R_{1-\delta} \mid i \in \mathcal{C}_k \cap \mathcal{A}_0^C\right) \geq 1 - \delta \quad \text{for each } k.$$

Under the sub-Gaussian assumption on $\boldsymbol{\epsilon}_i$, there exists a universal constant $c > 0$ such that

$$R_{1-\delta} \leq c\sigma\sqrt{d + \log(1/\delta)}.$$

Let the initialization error be $\Lambda_0 := \frac{1}{\Delta} \max_{k \in \{1,\ldots,K\}} \|\hat{\boldsymbol{\theta}}_k^{(0)} - \boldsymbol{\theta}_k\|$ and assume that

$$R_{1-\delta} < \left(\tfrac{1}{2} - \Lambda_0\right) \Delta. \tag{D.1}$$

Condition (D.1) implies that, with probability at least $1 - \delta$ for each $k$,

$$\|\boldsymbol{w}_i - \hat{\boldsymbol{\theta}}_k^{(0)}\| < \|\boldsymbol{w}_i - \hat{\boldsymbol{\theta}}_\ell^{(0)}\| \quad \text{for all } \ell \neq k, \quad i \in \mathcal{C}_k \cap \mathcal{A}_0^C. \tag{D.2}$$

In particular $\|\boldsymbol{w}_i - \hat{\boldsymbol{\theta}}_k^{(0)}\| < \tfrac{1}{2}\Delta$ and $\|\boldsymbol{w}_i - \hat{\boldsymbol{\theta}}_l^{(0)}\| \geq \tfrac{1}{2}\Delta$ for $l \neq k$.

Choose the trimming rate to cover true anomalies plus the inlier tail,

$$\tau \geq \varepsilon + \delta, \tag{D.3}$$

with $\varepsilon = N_a/N$. After the initial trimmed assignment that keeps the closest $(1-\tau)N$ observations to their nearest initialized center, all anomalous points are trimmed and at most a $\delta$ fraction of non-anomalous points are discarded. Hence the set passed to Lloyd's algorithm coincides with $\bigcup_k (\mathcal{C}_k \cap \mathcal{A}_0^C)$ up to at most $\delta N^\star$ losses. Denote $z_i = k$ as the true cluster assignment for each non-anomalous point $i \in \mathcal{C}_k \cap \mathcal{A}_0^C$, and let $\hat{z}_i^s$ be the assignment at iteration $s = 0, 1, 2, \ldots$. Define the mis-clustering rate on non-anomalous points as

$$A_s := \frac{1}{N^\star} \sum_{i \in \mathcal{A}_0^C} \mathbb{I}\{\hat{z}_i^s \neq z_i\}.$$

For the recovery guarantee, let $\pi_{\min} = \min_k \pi_k$ with $\pi_k = (N_k - N_k^a)/N^\star - \delta$ and

$$r_K = \frac{\Delta}{\sigma} \sqrt{\frac{\pi_{\min}}{1 + Kd/\{(1-\delta)N^\star\}}} \quad \text{and} \quad \Lambda_s = \frac{1}{\Delta} \max_{k \in \{1,\ldots,K\}} \|\hat{\boldsymbol{\theta}}_k^{(s)} - \boldsymbol{\theta}_k\|.$$

*Theorem D.1.* Assume $(1-\delta)N^\star \pi_{\min}^2 \geq CK \log\{(1-\delta)N^\star\}$ and $r_K \geq C\sqrt{K}$ for a sufficiently large constant $C$. Given any (data-dependent) initialization satisfying $\Lambda_0 \leq 1/2 - 4/\sqrt{r_K}$ with probability $1 - \nu$, we have

$$A_s \leq \exp\left(-\frac{\Delta^2}{16\sigma^2}\right) + \delta \quad \text{for all } s \geq 4\log N^\star$$

with probability at least $1 - \nu - 4/\{(1-\delta)N^\star\} - 2\exp(-\Delta/\sigma)$.

*Proof.* The result follows by applying Theorem 3.2 of Lu and Zhou [2016] to the retained observations obtained after the trimming step, under the stated initialization conditions. □

## D.2 Algorithm framework for Subject Segmentation and Cluster-Specific mFPCA

Algorithm 5 segments subjects into $K$ groups by combining their multivariate functional principal component (mFPC) embeddings with static covariates. First, for each subject $i$, the vector of its top $M$ mFPC scores is concatenated with its $q$-dimensional covariate vector to form a joint feature vector $\mathbf{w}_i \in \mathbb{R}^{M+q}$. The entire collection $\{\mathbf{w}_i\}_{i=1}^N$ is then mean–centered and whitened—i.e. multiplied by the inverse square root of its empirical covariance—so that the transformed features have zero mean and identity covariance. Finally, any off-the-shelf clustering algorithm (for example, K-means) is applied to the whitened feature set $\{\tilde{\mathbf{w}}_i\}$ to produce the final partition $\{\hat{\mathcal{C}}_k\}_{k=1}^K$.

---
**Algorithm 5** Clustering Subjects using mFPC Scores and Covariates
---
**Input:** mFPC scores $\{\hat{\rho}_{im}\}$ from Algorithm 1, and subject-level covariates $\{\mathbf{Z}_i \in \mathbb{R}^q\}$.
1: For $i = 1, \ldots, N$, get the mFPC score vector $\hat{\boldsymbol{\rho}}_i = (\hat{\rho}_{i1}, \ldots, \hat{\rho}_{iM})$ ▷ Algorithm 1.
2: Construct joint feature vectors $\mathbf{w}_i = (\hat{\boldsymbol{\rho}}_i, \mathbf{Z}_i) \in \mathbb{R}^{1 \times (M+q)}$ for $i = 1, \ldots, N$.
3: Standardize $\{\mathbf{w}_i\}$ to obtain whitened features $\{\tilde{\mathbf{w}}_i\}$: ▷ Standardisation
    a. Compute mean $\bar{\mathbf{w}} = \frac{1}{N}\sum \mathbf{w}_i$ and centered features $\mathbf{w}_{i,c} = \mathbf{w}_i - \bar{\mathbf{w}}$.
    b. Compute covariance $\hat{\boldsymbol{\Sigma}}_{\mathbf{w}} = \frac{1}{N-1}\sum \mathbf{w}_{i,c}^\top \mathbf{w}_{i,c}$ and its inverse square root $\hat{\boldsymbol{\Sigma}}_{\mathbf{w}}^{-1/2}$.
    c. $\tilde{\mathbf{w}}_i = \mathbf{w}_{i,c}\hat{\boldsymbol{\Sigma}}_{\mathbf{w}}^{-1/2}$. ▷ Features have zero mean, identity covariance
4: Apply trimmed $K$-means to $\{\tilde{\mathbf{w}}_i\}_{i=1}^N$ to obtain $K$ clusters $\{\hat{\mathcal{C}}_k\}_{k=1}^K$.
**Output:** A set of $K$ clusters $\{\hat{\mathcal{C}}_k\}_{k=1}^K$.
---

Within each cluster $\mathcal{C}_\ell$, we re-apply the Multivariate Sparse FPCA procedure to obtain cluster-specific mean functions $\hat{\mu}_{\mathcal{C}_\ell}^{(d)}(t)$, eigenfunctions $\hat{\psi}_{m,\mathcal{C}_\ell}^{(d)}(t)$, and updated scores $\hat{\rho}_{im,\mathcal{C}_\ell}$ using algorithm 6 below.

Algorithm 6 revisits each cluster identified in Algorithm 5 and applies the multivariate sparse FPCA pipeline to the subjects within that cluster. For each cluster $\hat{\mathcal{C}}_k$, it collects the raw SL trajectories, invokes the mFPCA routine (Algorithm 1), and produces cluster-specific FPC scores, eigenfunctions, and mean curves that capture the principal modes of joint variation within that group.

---

**Algorithm 6** Cluster-Specific mFPCA

---

**Input:** Data $\{\boldsymbol{Y}_i(T_{ij})\}_{i=1}^N$, and the set of $K$ clusters $\{\hat{\mathcal{C}}_k\}_{k=1}^K$ from Algorithm 5.

1: **for** $k = 1, \ldots, K$ **do**  ▷ For each cluster
2:  Let $\mathcal{Y}_{\hat{\mathcal{C}}_k} = \{\boldsymbol{Y}_i(T_{ij}) : i \in \hat{\mathcal{C}}_k\}$.  ▷ Data for subjects in cluster $\hat{\mathcal{C}}_k$
3:  $(\{\hat{\rho}_{im}^k\}_{i \in \hat{\mathcal{C}}_k}, \{\hat{\boldsymbol{\psi}}_m^k\}, \{\hat{\boldsymbol{\mu}}_k\}) \leftarrow \text{mFPCA}(\mathcal{Y}_{\hat{\mathcal{C}}_k})$  ▷ Algorithm 1
4: **end for**

**Output:** For each cluster $\hat{\mathcal{C}}_k$ ($k = 1, \ldots, K$): cluster-specific mFPC scores $\{\hat{\rho}_{im}^k\}_{i \in \hat{\mathcal{C}}_k}$, mean function $\{\hat{\boldsymbol{\mu}}_k(t)\}$, and eigenfunctions $\{\hat{\boldsymbol{\psi}}_m^k\}$.

---

The final subject-specific $p$-variate trajectory, incorporating cluster-specific structure, $\hat{\boldsymbol{X}}_{i,\hat{\mathcal{C}}_\ell}(t)$ for subject $i \in \hat{\mathcal{C}}_\ell$, is then given by:

$$\widehat{X}_{i,\mathcal{C}_\ell}^{(d)}(t) = \hat{\mu}_{\mathcal{C}_\ell}^{(d)}(t) + \sum_{m=1}^M \hat{\rho}_{im,\mathcal{C}_\ell} \, \hat{\psi}_{m,\mathcal{C}_\ell}^{(d)}(t), \quad d = 1, \ldots, p, \tag{D.4}$$

thereby aligning each subject's fitted trajectory with the structural characteristics of its assigned cluster.

## D.3 Cluster Stability

Though clustering is fully unsupervised and lacks ground truth labels, several validation metrics exist to assess cluster quality. In our implementation of LLmFPCA-detect across both domains, we selected the optimal number of clusters based on the average silhouette width. LLmFPCA-detect can, in general, accommodate any clustering method without altering the overall workflow. For the Amazon Reviews dataset, we assessed cluster stability using the bootstrapped Jaccard index, a common metric in the clustering literature. Across 200 bootstrap samples, the mean Jaccard indices for Clusters 1–3 were 77%, 91%, and 76%, respectively. The scores being $\geq 75\%$ confirm the reproducibility and stability of the clusters. A similar validation was performed on the Wikipedia dataset, yielding 94% for Cluster 1 and 87% for Cluster 2, further demonstrating consistent cluster recovery under resampling.

# E    Principles of Anomaly Detection with Subroutine Details for Algorithms 2 and 3 in Section 3

## E.1    Principles of Anomaly Detection

For $i \in \mathcal{A}_0$ suppose that the random variable $\langle \boldsymbol{a}_i, \boldsymbol{\psi}_m \rangle$ is distributed according to a continuous law $\mathcal{L}_{m,a}$ with compact support $[a_{min}, a_{max}]$ where $a_{min}, a_{max} > 0$. Once the clusters have been obtained, one can extract the cluster-specific mFPCA of the trajectories and recompute the mFPCA scores as

$$\rho_{im}^k = \langle \boldsymbol{Y}_i - \boldsymbol{\mu}_k, \boldsymbol{\psi}_m \rangle = \langle \boldsymbol{X}_i - \boldsymbol{\mu}_k, \boldsymbol{\psi}_m \rangle + \langle \boldsymbol{\eta}_i, \boldsymbol{\psi}_m \rangle + \alpha_{im}$$

where $\alpha_{im} = \langle \boldsymbol{a}_i, \boldsymbol{\psi}_m \rangle$. For $i \in C_k \cap \mathcal{A}_0^C$, define $\mathcal{L}_m^k$ as the law of $\rho_{im}^k = \langle \boldsymbol{Y}_i - \boldsymbol{\mu}_k, \boldsymbol{\psi}_m \rangle = \langle \boldsymbol{X}_i - \boldsymbol{\mu}_k, \boldsymbol{\psi}_m \rangle + \langle \boldsymbol{\eta}_i, \boldsymbol{\psi}_m \rangle$ as $\alpha_{im} = 0$ in this case. Let $\rho_m^k \sim \mathcal{L}_m^k$, and assume that $\rho_m^k$ has a continuous distribution. For the anomalous points $i \in C_k \cap \mathcal{A}_0$, the scores $\rho_{im}^k = \langle \boldsymbol{Y}_i - \boldsymbol{\mu}_k, \boldsymbol{\psi}_m \rangle = \langle \boldsymbol{X}_i - \boldsymbol{\mu}_k, \boldsymbol{\psi}_m \rangle + \langle \boldsymbol{\eta}_i, \boldsymbol{\psi}_m \rangle + \alpha_{im}$ are assumed to be independent, with $\alpha_{im} \overset{iid}{\sim} \mathcal{L}_{m,a}$. Thus, the law of $\rho_{im}^k$ corresponds to that of $\rho_m^{a,k} = \rho_m^k + \alpha_m$, where $\alpha_m \sim \mathcal{L}_{m,a}$ is independent of $\rho_m^k$. Let $\tilde{\rho}_m^k$ be a random variable drawn from the empirical distribution of $\rho_{im}^k$ over $i \in C_k$. For any given $\epsilon > 0$, define the set $\mathcal{A}_0^{k,\epsilon}$ as $\mathcal{A}_0^{k,\epsilon} = \{i \in C_k : \mathbb{P}(\bigcup_{m=1}^M \{|\tilde{\rho}_m^k| > |\rho_{im}^k|\}) < \epsilon\}$. Theorem E.1 states that one can screen for the anomalous points in $C_k$ by using the tail of the distribution of $\tilde{\rho}_m^k$. While this may not recover $C_k \cap \mathcal{A}_0$ exactly, observe that for $i \in \mathcal{A}_0^{k,\epsilon} \setminus \{C_k \cap \mathcal{A}_0\}$, $i \in C_k \cap \mathcal{A}_0^C$ and therefore $\alpha_{im} = 0$ which implies that $\mathbb{P}(|\rho_m^k| > |\rho_{im}^k|) = 0.5$ for any $m \in \{1, \ldots, M\}$. Hence $\mathbb{P}(\bigcup_{m=1}^M |\rho_m^k| > |\rho_{im}^k|) \geq 0.5$, which inspires the calibration step of the screened points in $\mathcal{A}_0^{k,\epsilon}$ to recover the set $\{C_k \cap \mathcal{A}_0\}$ accurately.

*Theorem E.1.* Assume that for any $k = 1, \ldots, K$, $\pi_{a,k} = o(1)$ as $N \to \infty$. Then, there exists $\epsilon > 0$

such that for any $k \in \{1, \ldots, K\}$, $C_k \cap \mathcal{A}_0 \subset \mathcal{A}_0^{k,\epsilon}$.

*Proof.* Let $i \in C_k \cap \mathcal{A}_0$. Observe that

$$\mathbb{P}\left(\bigcup_{m=1}^{M}\{|\tilde{\rho}_m^k| > |\rho_{im}^k|\}\right)$$
$$\leq \sum_{m=1}^{M} \mathbb{P}(|\tilde{\rho}_m^k| > |\rho_{im}^k|)$$
$$= \sum_{m=1}^{M} \left\{(1 - \pi_{a,k})\mathbb{P}(|\rho_m^k| > |\rho_{im}^k|) + \pi_{a,k}\mathbb{P}(|\rho_m^{a,k}| > |\rho_{im}^k|)\right\}$$
$$\leq (1 - \pi_{a,k}) \sum_{m=1}^{M} \mathbb{P}(|\rho_m^k| > |\rho_{m,2}^k + a_{min}|) + \pi_{a,k} M/2$$

where the second term follows as for any $i \in C_k \cap \mathcal{A}_0$, $\rho_m^{a,k}$ and $\rho_{im}^k$ are identically distributed and for the first term, observe that for any $i \in C_k \cap \mathcal{A}_0$, $\mathbb{P}(|\rho_m^k| > |\rho_{im}^k|) \leq \mathbb{P}(|\rho_m^k| > |\rho_{m,2}^k + a_{min}|)$ with $\rho_{m,2}^k$ being an i.i.d copy of $\rho_m^k$ as $\rho_{im}^k$ has the same distribution as $\rho_{m,2}^k + \alpha_m$ and $\alpha_m \geq a_{min}$ almost surely. Let $\tilde{\epsilon} = \max_{m \in 1, \ldots, M} \left\{\mathbb{P}(|\rho_m^k| > |\rho_{m,2}^k + a_{min}|)\right\} > 0$. Then,

$$\mathbb{P}\left(\bigcup_{m=1}^{M}\{|\tilde{\rho}_m^k| > |\rho_{im}^k|\}\right) \leq (1 + M)\tilde{\epsilon}$$

which completes the proof by taking $\epsilon = (1 + M)\tilde{\epsilon}$. □

## E.2 Subroutine Details for Algorithm 2

Subroutine algorithm 7 (`ScreenPotentialOutliers`): This routine identifies subjects whose multivariate functional principal component (mFPC) scores lie in the extreme tails of their cluster-specific distribution. Given a set of subjects $I_s$ and their scores on $B$ mFPC components, we first build, for each component $m$, the empirical cumulative distribution function $\widehat{F}_{m,I_s}$. We then extract the lower and upper cutoff points at probabilities $\alpha_1/(2B)$ and $1 - \alpha_1/(2B)$, respectively. A subject $i$ is flagged as a potential outlier if any of its component scores falls below the lower cutoff or above the upper cutoff. All such flagged indices comprise the set $G_s$, while the remaining "clean" subjects form $G_s^c = I_s \setminus G_s$. By focusing on extreme quantiles of each component's score distribution, this screening step ensures that only those subjects with unusually large or small loadings proceed to the subsequent confirmation stage.

---

**Algorithm 7** `ScreenPotentialOutliers`: Screening for Potential Outliers in a Subject Set

---

**Input:** Set of subject indices $I_s \subset \hat{\mathcal{C}}$; cluster-specific mFPC scores $\{\hat{\rho}_{jm,\hat{\mathcal{C}}} : j \in I_s, m = 1, \ldots, B\}$; number of components $B$; screening level $\alpha_1$.

1: Initialize $G_s \leftarrow \emptyset$.     ▷ Potential outliers in $I_s$
2: **for** $m = 1, \ldots, B$ **do**
3:     Let $\widehat{F}_{m,I_s}$ be the empirical CDF of scores $\{\hat{\rho}_{jm,\hat{c}} : j \in I_s\}$.
4:     $q^{\text{lo}}_{m,I_s} \leftarrow \widehat{F}^{-1}_{m,I_s}(\alpha_1/(2B))$; $q^{\text{hi}}_{m,I_s} \leftarrow \widehat{F}^{-1}_{m,I_s}(1 - \alpha_1/(2B))$.     ▷ Empirical quantiles
5: **end for**
6: **for** each subject $i \in I_s$ **do**
7:     **if** $\exists m \in \{1, \ldots, B\}$ s.t. $(\hat{\rho}_{im,\hat{c}} < q^{\text{lo}}_{m,I_s}$ or $\hat{\rho}_{im,\hat{c}} > q^{\text{hi}}_{m,I_s})$ **then**
8:         Add $i$ to $G_s$.
9:     **end if**
10: **end for**
11: $G_s^c \leftarrow I_s \setminus G_s$.     ▷ Clean held-out set from $I_s$

**Output:** Set of potential outliers $G_s$; clean held-out set $G_s^c$.

---

Subroutine algorithm 8 (`ConfirmAnomalies`): This routine takes as input a candidate set of subjects $I_{\text{test}}$, a calibration set $I_{\text{calib}}$, cluster-specific mFPC scores for $B$ components, and a global significance level $\alpha$. Its goal is to confirm which candidates exhibit unusually large scores relative to the calibration group.

For each subject $i \in I_{\text{test}}$, we initialize an empty index set $S_i$ to record the components in which $i$ may be anomalous. For each component $m = 1, \ldots, B$, we compute an empirical $p$-value by counting how many calibration scores exceed $|\hat{\rho}_{im}|$, adding one to both numerator and denominator to guard against zero counts. We then compare this $p$-value to the adjusted threshold $\alpha/B$; if $p < \alpha/B$, component $m$ is flagged and added to $S_i$. After all components are tested, any subject with $S_i \neq \emptyset$ is declared a confirmed anomaly, and the pair $(i, S_i)$ is appended to the output set $\mathcal{A}_{\text{confirmed}}$.

The subroutine returns $\mathcal{A}_{\text{confirmed}}$, the list of all subjects whose functional scores deviate significantly from the calibration distribution, together with the specific components in which each deviation occurs.

**Algorithm 8** ConfirmAnomalies: Confirming Anomalous Subjects using Calibration

**Input:** Set of potential outlier indices $I_{test}$; calibration set $I_{calib}$; cluster-specific mFPC scores $\{\hat{\rho}_{jm,\hat{\mathcal{C}}} : j \in I_{test} \cup I_{calib}, m = 1, \ldots, B\}$; number of components $B$; significance level $\alpha$.

1: Initialize $\mathcal{A}_{confirmed} \leftarrow \emptyset$.
2: **for** each subject $i \in I_{test}$ **do**
3:    $S_i \leftarrow \emptyset$.                                                                  ▷ Captures outlying FPC components for subject $i$
4:    **for** $m = 1, \ldots, B$ **do**
5:       $p_{im}^{\text{emp}} \leftarrow \frac{1 + \#\{j \in I_{calib} : |\hat{\rho}_{jm,\hat{\mathcal{C}}}| \geq |\hat{\rho}_{im,\hat{\mathcal{C}}}|\}}{1 + |I_{calib}|}$.       ▷ Compute empirical p-value
6:       **if** $p_{im}^{\text{emp}} < \alpha/B$ **then**       ▷ Bonferroni correction; other testing methods (e.g., BH) applicable
7:          Add $m$ to $S_i$.
8:       **end if**
9:    **end for**
10:   **if** $S_i \neq \emptyset$ **then**
11:      Add $(i, S_i)$ to $\mathcal{A}_{confirmed}$.
12:   **end if**
13: **end for**

**Output:** Set of confirmed anomalous subjects from $I_{test}$ with their deviating components $\mathcal{A}_{confirmed} = \{(i, S_i)\}$.

### E.3 Subroutine Details for Algorithm 3

Subroutine algorithm 9 (`ComputeWindowDeviations`): This routine prepares the baseline deviation measures that will later be used to identify time-localized anomalies. It accepts as input the observed trajectories of all calibration subjects, the estimated cluster-level mean functions, and a predefined collection of time windows.

For each window, the algorithm first computes the average value of the cluster mean over that interval. It then, for each calibration subject, calculates the subject's own average observation in the same interval and measures the deviation as the maximum absolute difference between the subject's window average and the cluster mean average. Each deviation score is recorded in a lookup table indexed by subject and window.

On completion, the subroutine outputs two sets of results: (1) the window-specific cluster averages and (2) the matrix of deviation scores for every calibration subject across all windows. These precomputed quantities serve as the reference distribution when testing whether any subject's windowed behavior departs significantly from the cohort norm.

Subroutine algorithm 10 (`IdentifyAnomalousWindows`): This routine takes the set of subjects

**Algorithm 9** ComputeWindowDeviations: Precompute Window Deviations for Calibration

**Input:** Data $\{\boldsymbol{Y}_j(T_{jk}) : j \in G_1^c \cup G_2^c\}$ for calibration subjects; cluster means $\{\hat{\mu}_{\hat{\mathcal{C}}}^{(d)}(t)\}$; time windows $\{(a_w, b_w)\}_{w=1}^{W}$.

1: Initialize map $D_{calib} \leftarrow \emptyset$ for storing $D_j^{(w)}$ values.
2: Initialize map $\overline{\mathcal{M}}_{\mathcal{C}} \leftarrow \emptyset$ for storing $\bar{\boldsymbol{\mu}}_{\hat{\mathcal{C}}}^{(w)}$ values.
3: **for** each window $w = 1, \ldots, W$ **do**
4: $\quad \bar{\boldsymbol{\mu}}_{\hat{\mathcal{C}}}^{(w)} \leftarrow (\text{avg}(\hat{\mu}_{\hat{\mathcal{C}}}^{(1)}(t) \text{ in } (a_w, b_w]), \ldots, \text{avg}(\hat{\mu}_{\hat{\mathcal{C}}}^{(p)}(t) \text{ in } (a_w, b_w]))$.
5: $\quad$ Store $\bar{\boldsymbol{\mu}}_{\hat{\mathcal{C}}}^{(w)}$ in $\overline{\mathcal{M}}_{\mathcal{C}}$ indexed by $w$.
6: $\quad$ **for** each subject $j \in G_1^c \cup G_2^c$ **do**
7: $\quad\quad \bar{\mathbf{e}}_j^{(w)} \leftarrow$ average of $\{\boldsymbol{Y}_j(T_{jk}) \text{ where } T_{jk} \in (a_w, b_w]\}$.
8: $\quad\quad D_j^{(w)} \leftarrow \|\bar{\mathbf{e}}_j^{(w)} - \bar{\boldsymbol{\mu}}_{\hat{\mathcal{C}}}^{(w)}\|_{\infty}$.
9: $\quad\quad$ Store $D_j^{(w)}$ in $D_{calib}$ indexed by $(j, w)$.
10: $\quad$ **end for**
11: **end for**

**Output:** Windowed cluster means $\{\bar{\boldsymbol{\mu}}_{\hat{\mathcal{C}}}^{(w)}\}_{w=1}^{W}$ (via map $\overline{\mathcal{M}}_{\mathcal{C}}$); Calibration deviation scores $\{D_j^{(w)}\}_{j \in G_1^c \cup G_2^c, w=1,\ldots,W}$ (via map $D_{calib}$).

already confirmed as anomalous, their observed trajectories, the precomputed windowed cluster means, and the calibration deviation scores from the two held-out groups. For each flagged subject, we choose the calibration group that does not include that subject. We then break the subject's trajectory into $W$ contiguous time windows. Within each window, we compute the subject's average observation and measure its maximum absolute difference from the corresponding cluster mean. This difference is compared against the calibration scores for that window to form an empirical $p$-value—namely, the fraction of calibration subjects whose deviation equals or exceeds the subject's own, with a small offset in numerator and denominator to avoid zero counts. After applying a Bonferroni correction across all $W$ windows, any window with $p < \alpha/W$ is marked as anomalous. The subroutine outputs the final list of subjects together with the specific windows in which each one deviates significantly from the cohort norm.

# F Supplemental details for Dynamic Keyword Profiling

To extract intent keywords from anomalous reviews, we develop a dynamic and scalable algorithm that leverages a large language model (LLM) to identify and update user intent lists over time.

---
**Algorithm 10** `IdentifyAnomalousWindows`: Identify Anomalous Windows for Subjects
---
**Input:** Anomalous subjects $\mathcal{A}^{(1)}$; data $\{\boldsymbol{Y}_i(T_{ij}) : i \text{ s.t. } (i, \_) \in \mathcal{A}^{(1)}\}$; windowed cluster means $\{\bar{\boldsymbol{\mu}}_{\hat{\mathcal{C}}}^{(w)}\}$; calibration deviation scores $\{D_j^{(w)}\}$; calibration split info $I_1, I_2, G_1^c, G_2^c$; time windows $\{(a_w, b_w]\}_{w=1}^W$; significance level $\alpha$.

1: Initialize $\mathcal{A}^{(2)} \leftarrow \emptyset$.
2: **for** each subject $i$ such that $(i, \_) \in \mathcal{A}^{(1)}$ **do**
3:     Determine $G_{\text{null}}^c$: if $i \in I_1$, $G_{\text{null}}^c \leftarrow G_2^c$; else $G_{\text{null}}^c \leftarrow G_1^c$.
4:     Initialize $\mathcal{W}_i \leftarrow \emptyset$.      ▷ Anomalous windows for subject $i$
5:     **for** each window $w = 1, \ldots, W$ **do**
6:         $\bar{\mathbf{e}}_i^{(w)} \leftarrow$ average of $\{\boldsymbol{Y}_i(T_{ij}) \text{ where } T_{ij} \in (a_w, b_w]\}$.
7:         $D_i^{(w)} \leftarrow \|\bar{\mathbf{e}}_i^{(w)} - \bar{\boldsymbol{\mu}}_{\hat{\mathcal{C}}}^{(w)}\|_\infty$.
8:         $p_i^{(w)} \leftarrow \frac{1 + \#\{j \in G_{\text{null}}^c : D_j^{(w)} \geq D_i^{(w)}\}}{1 + |G_{\text{null}}^c|}$.      ▷ Uses precomputed $D_j^{(w)}$
9:         **if** $p_i^{(w)} < \alpha/W$ **then**      ▷ Bonferroni correction
10:             Add $w$ to $\mathcal{W}_i$.
11:         **end if**
12:     **end for**
13:     **if** $\mathcal{W}_i \neq \emptyset$ **then**
14:         Add $(i, \mathcal{W}_i)$ to $\mathcal{A}^{(2)}$.
15:     **end if**
16: **end for**

**Output:** Set $\mathcal{A}^{(2)} = \{(i, \mathcal{W}_i) : i \in \mathcal{A}^{(1)}, \mathcal{W}_i \neq \emptyset\}$.
---

The goal is to cluster semantically similar review content into concise intent phrases, even when expressed using varied vocabulary and writing styles.

Each user $j \in \mathcal{U}$ is associated with a set of anomalous reviews $S_j$, where each review may include a descriptive title. At any time $t$, we maintain a user-specific, time-ordered sequence of extracted intents $I_j^{(t-)}$ representing the intents derived from user $j$'s prior reviews. For each new review $r_j^{(t)}$, the algorithm compares its content to the top $k$ prior intents for user $j$, denoted $\mathcal{K}_j^{(t-)} = \text{Top}_k(I_j^{(t-)})$, as well as the top $k$ intents across all other users, $\mathcal{K}_{-j}^{(t-)} = \bigcup_{l \neq j} \mathcal{K}_l^{(t-)}$.

The combined set $\{I_j^{(t-)}, \mathcal{K}_{-j}^{(t-)}, r_j^{(t)}\}$ is fed into an LLM. If the LLM determines that the review matches an existing intent from either the user's own history or from others, it is assigned to that intent. Otherwise, the model generates a new intent phrase tailored to the content of the review, which is then appended to $I_j^{(t)}$. This process is outlined formally below.

**Algorithm 11** Intent Keyword Extraction from Anomalous Reviews

**Input:**

- Users $\mathcal{U}$ and their anomalous reviews $S_j$.
- For each user $j$: prior intent list $I_j^{(t-)}$.
- For each time $t$: new review $r_j^{(t)}$.
- Top $k$ intents from user $j$: $\mathcal{K}_j^{(t-)}$.
- Top $k$ intents from others: $\mathcal{K}_{-j}^{(t-)}$.

1: **for** each user $j \in \mathcal{U}$ **do**
2:     **for** each time $t$ with anomalous review $r_j^{(t)}$ **do**
3:         Retrieve $I_j^{(t-)}$
4:         Compute $\mathcal{K}_j^{(t-)} = \text{Top}_k(I_j^{(t-)})$
5:         Compute $\mathcal{K}_{-j}^{(t-)} = \bigcup_{l \neq j} \mathcal{K}_l^{(t-)}$
6:         Query LLM with $\{I_j^{(t-)}, \mathcal{K}_{-j}^{(t-)}, r_j^{(t)}\}$
7:         **if** LLM matches $r_j^{(t)}$ to an intent $i^* \in I_j^{(t-)} \cup \mathcal{K}_{-j}^{(t-)}$ **then**
8:             Assign $r_j^{(t)}$ to $i^*$
9:         **else**
10:            Generate new intent $i_{\text{new}} = \text{LLM}(r_j^{(t)})$
11:            Update $I_j^{(t)} \leftarrow I_j^{(t-)} \cup \{i_{\text{new}}\}$
12:         **end if**
13:         Update $\mathcal{K}_j^{(t)} = \text{Top}_k(I_j^{(t)})$
14:     **end for**
15: **end for**

**Output:**

- Updated time-ordered intent lists $I_j$ for each user.
- Top $k$ intent keywords $\mathcal{K}_j$ for each user.
- Mapping from reviews to associated intent keywords.

To illustrate the types of intents extracted and their temporal distribution, Table 6 summarizes keyword frequencies by cluster across four time windows. Clusters reflect coherent intent themes—such as product quality, value for money, or fit—and demonstrate the algorithm's ability to maintain semantic consistency while adapting to newly emerging patterns. See Table 6 in Section A for details.

# G  Emotion scoring and validation

Emotion labels are generated for each Amazon review using GPT-4-32k according to Plutchik's wheel (see Prompt 2). To validate our embedding, we selected 30 reviews and had humans annotate each on the 24 primary and opposing petals. We then applied the same GPT prompt (Prompt 2) to these examples and compared its output to the human consensus. The exact-match accuracy—defined as the fraction of reviews where GPT's predicted emotion agreed with the expert label—was 19/33 (57.6%).

**Listing 2:** *Prompt for Emotion Category Classification*

```
--- CONTEXT ---
You are an expert at honestly classifying emotion from text. This task involves
   analyzing user reviews of products, each consisting of a review_title and
   review_text. The objective is to identify the predominant emotion expressed based on
    a structured framework based on 8 main emotions, each with three levels of
   intensity, leading to a total of 24 distinct emotional categories.

--- INSTRUCTIONS ---
Carefully read each user review, including both the review_title and review_text. Based
   on the emotional cues present,
use the structured framework of 8 primary emotions, described below, along with their
   respective intensities, to guide your analysis,
and classify the review into one of the 24 distinct emotional categories.

1. Joy: Indicates happiness or pleasure derived from product satisfaction.
   - Serenity (Mild)
   - Joy (Moderate)
   - Ecstasy (Intense)

2. Trust: Suggests reliability or confidence in the product.
   - Acceptance (Mild)
```

- Trust (Moderate)

   - Admiration (Intense)

3. Fear: Shows worry or concern about the product or its effects.

   - Apprehension (Mild)

   - Fear (Moderate)

   - Terror (Intense)

4. Surprise: Reflects astonishment or unexpected reactions towards the product.

   - Distraction (Mild)

   - Surprise (Moderate)

   - Amazement (Intense)

5. Sadness: Reveals disappointment or sorrow due to unmet product expectations.

   - Pensiveness (Mild)

   - Sadness (Moderate)

   - Grief (Intense)

6. Disgust: Demonstrates revulsion or strong disapproval of the product.

   - Boredom (Mild)

   - Disgust (Moderate)

   - Loathing (Intense)

7. Anger: Exhibits frustration or anger towards the product or service.

   - Annoyance (Mild)

   - Anger (Moderate)

   - Rage (Intense)

8. Anticipation: Expresses hopeful expectation or eagerness about the product.

   - Interest (Mild)

```
    - Anticipation (Moderate)
    - Vigilance (Intense)

--- TASK ---
For each user review_title and review_text, follow the instructions to categorize the
    emotion expressed in the review
into one of the 24 emotion categories. Ensure your classification is presented as a
    single word in Python string format.
```

Below we present the table of human annotations and GPT-predicted Plutchik dimensions for a set of representative Amazon reviews. Experts labeled each review according to one of the twenty four possible emotion categories as per the eight primary petals of Plutchik's Wheel (four opposing emotion pairs), and we then applied our GPT-4 prompt (Listing 2) to generate a corresponding prediction. Table 8 showcases these results, highlighting both concordant and discordant cases, which form the basis for our exact-match accuracy metric reported in Section 4.1.

Table 8: Comparison of human-annotated vs. GPT-predicted Plutchik dimensions.

| Title | Review text | Emotion (human) | Emotion (GPT) |
| --- | --- | --- | --- |
| Perfect Fit | for Polaris Product as advertised, and it fit perfectly in my Polaris ATV. | joy | Joy |
| Good filter wrench | Works great for Mercedes V6 Diesel engine oil change. Matter of fact, can't change the oil filter without it. Bit pricey for what it is, but works well. | acceptance/submission | Trust |

*(continued on next page)*

| Title | Review text | Emotion (human) | Emotion (GPT) |
| --- | --- | --- | --- |
| Same as name brand but cost less | Great aftermarket and non-name-brand insect screen for your RV furnace exhaust. Same product that comes in fancy packaging, but less expensive. I would buy this again. | joy | Trust |
| 100% lambs wool | Product does an OK job. Not sure if it's that much better than a synthetic cleaning tool for an RV. Clearly not sure about value received for dollars spent. Will have to use more as time progresses. Get the smaller unit as the bigger one is probably way too hard. | apprehension | Apprehension |
| Easy to use | Easy to see waste water from holding tanks so you know when they are empty and clean, mainly black water tank. | joy | Trust |
| Good value | Decently made and works ok. | acceptance | Acceptance |
| Well made product | Really happy with this purchase to use with our motorhome. Just the right height at lowest level and has many more adjustments if needed for uneven ground. It is well made, all the welds are clean and even, and it has non-slip tape attached. I would recommend. | joy | Joy |

| Title | Review text | Emotion (human) | Emotion (GPT) |
| --- | --- | --- | --- |
| Great customer service | Purchased this to use for my motorhome and the 22.5″ tires that are running at 100 psi. The unit is quiet, well made, with the exception of the plastic screw-on cap for the filter as it is cast plastic. I did have a problem with the air chuck that came. | trust | Trust |
| Simple to install | Easy to install and setup, once you watch the videos made by TST. We use this in our Class A motorhome and towed vehicle. The color display/monitor is nice and easy to see and read. TST provides both a movable dash mount or window suction mount. | admiration | Trust |
| Does the Job | Just a gas can that has been engineered to prevent dummies from spilling gas. | acceptance | Acceptance |
| OEM oil filter | You'd think Mercedes would have moved to a canister style filter rather than the old school paper. But it works. Used on a MBS Sprinter Class C motorhome. | acceptance | Acceptance |
| OEM Hydraulic Oil | Used as a backup for my HWH hydraulic system on my motorhome. Good, OEM product. | trust | Trust |

| Title | Review text | Emotion (human) | Emotion (GPT) |
| --- | --- | --- | --- |
| OEM part | Great price on OEM part for my car. Works great. Very happy with part. | joy | Joy |
| Good replacement item | Great price on air filter for car. Half the price that shop wanted for air filter. | joy | Joy |
| Appear wellmade | Having just received these, and confirmed they do fit my 22.5″ tires on my class A, I would have to say that these appear to be a well made product. They are easy to install and I appreciate the retainer strap that goes behind the tire to secure the cover. | trust | Trust |
| Genuine parts | Basic Onan generator filter. | acceptance | Acceptance |
| Do the job, at a cost | These crush washers work for the oil drain plug on my Subaru but was surprised at the cost for two. | acceptance | Surprise |

*(continued on next page)*

| Title | Review text | Emotion (human) | Emotion (GPT) |
| --- | --- | --- | --- |
| Simple to use | Seems an accurate device as compared to the Milton gauge I have been using. On my motorhome, this fits the valve stems better than the Milton, thereby not losing air pressure while reading tire pressure. Overall, happy with the device. | joy | Joy |
| It's a fuse | Priced well and hope I don't have to use it. | acceptance | Anticipation |
| Good fit for generator | Appears well made and machined. Will fit an Onan 8000 KW diesel generator in our motorhome. For the price, it should also come with the plastic cover to keep debris out of the drain hole. | acceptance | Acceptance |
| Good purchase | Perfect fit for Onan QD8000 generator and priced well. | joy | Trust |
| Decent Wax | It's a wax, it works. | acceptance | Acceptance |
| Works OK | Decent wax for boats. | acceptance | Acceptance |
| Good size | Works well for Diesel engine oil changes. | joy | Trust |

| Title | Review text | Emotion (human) | Emotion (GPT) |
| --- | --- | --- | --- |
| Good towels | Work well for use on motorhome windows or waxing. | joy | Acceptance |
| Decent wash nut | Does an OK job of washing vehicles without abrasive issues. | acceptance | Acceptance |
| Direct Replacement | This light fixture is a direct replacement for my lights in my motorhome closets and storage bays. | acceptance | Acceptance |
| Absolutely Junk! Do not buy | Received this today and went to put it on my 3/8″ extension for an oil filter change. The machining is pretty, but measurements are so poor I cannot get it on the extension to use. Absolute junk! I should have paid more attention to the negative review. | rage | Rage |
| OEM Filter | Always want to use the OEM filter on my PSD. | trust | Trust |
| Simple Installation | Easy to install, and should be a breeze for cleaner drains. The only downside I can see is that it will take longer to drain and will require staying under the truck to get to second container for final drain. I installed this on an F350 with a 6.4 L PSD. | interest | Trust |

| Title | Review text | Emotion (human) | Emotion (GPT) |
| --- | --- | --- | --- |
| Good Product for RV use | Product appears to be well made if brass and stainless steel. Just using it now for the first time and it is holding water pressure consistently at the campsite. Makes me wonder why I waited so long to upgrade my old, in-line regulator, which will now be. | joy | Trust |
| Simplifies oil change | Simple to install. Does take longer to drain, as it is a smaller opening the oil drains through. I use an adapter and tube to drain into 1 gallon jugs, so no real cleanup involved. | acceptance | Trust |
| Reliable product | Works well to ensure toilet seal remains pliable. | acceptance | Trust |

# H  LLM usage

In preparing this manuscript, we employed large language model (LLM) solely as an assistive tool to aid in polishing and refining the writing. Specifically, LLM was used to improve the grammar, clarity, and readability of the text.

# I  Table of Notations

Additionally Table 9 compiles all key symbols and their definitions for easy reference. Each entry lists the mathematical notation, its interpretation in the context of our customer journey analysis, and the section where it is first introduced. This comprehensive glossary ensures clarity and consistency across the various methodological components described in this paper.

Table 9: Table for a complete list of symbols and their meanings as used in this paper.

| Notation | Interpretation | Location defined |
| --- | --- | --- |
| $i \in \{1, \ldots, N\}$ | Subject index, total of $N$ users | Sec. 3 |
| $N_i$ | Number of observations for subject $i$ | Sec. 3 |
| $T_{ij}$ | Observation time of the $j$th record for user $i$ | Sec. 3 |
| $K_i(T_{ij})$ | Raw text transcript at time $T_{ij}$ | Sec. 3 |
| $\Phi : \mathcal{X} \to \mathbb{R}^p$ | Embedding map from text to $p$-dim numeric vector | Sec. 3 |
| $\boldsymbol{Y}_i(T_{ij})$ | $p$-variate embedding at time $T_{ij}$ | Sec. 3 |
| $\hat{\mu}^{(d)}(t),\ \hat{\phi}_k^{(d)}(t),\ \hat{\xi}_{ik}^{(d)}$ | Univariate mean, eigenfunctions, and scores | Alg. 4 |
| $\hat{\Xi} \in \mathbb{R}^{N \times M}$ | Stacked univariate FPC scores | Alg. 1 |
| $\hat{\lambda}_m,\ \hat{v}_m$ | Eigenvalues and eigenvectors of $\hat{C}_\Xi$ | Alg. 1 |
| $\hat{\psi}_m^{(d)}(t)$ | Multivariate eigenfunction, component $m$, dimension $d$ | Alg. 1 |
| $\hat{\rho}_{im}$ | mFPCA score for user $i$, component $m$ | Alg. 1 |
| $\widehat{X}_{i,\mathcal{C}_\ell}^{(d)}(t)$ | Reconstructed cluster specific $p$-variate trajectory | Sec. D.2 |
| $\hat{\mathcal{C}}_\ell$ | Cluster $\ell$ via K-means | Sec. D |
| $I_1, I_2$ | Random half-splits for screening | Alg. 2 |
| $\alpha_1, \alpha$ | Screening and final significance levels | Alg. 2 |
| $\mathcal{A}^{(1)}$ | Confirmed anomalies w.r.t. cohort | Alg. 2 |
| $(a_w, b_w]$ | Fixed time window $w$ | Alg. 3 |
| $D_j^{(w)}$ | Deviation score for user $j$ in window $w$ | Alg. 9 |
| $\mathcal{A}^{(2)}$ | Confirmed anomalous windows | Alg. 3 |
| $\Delta$ | Minimum separation between true cluster centroids | App. D (Thm. D.1) |
| $\pi_k = N_k/N$ | Proportion of cluster $\mathcal{C}_k$ | App. D (Thm. D.1) |
| $\pi_{\min} = \min_k \pi_k$ | Smallest cluster proportion | App. D (Thm. D.1) |
| $N_a, N^\star$ | Total anomalies, total non-anomalous | App. D (Thm. D.1) |
| $r_k$ | Normalized SNR for cluster $k$ | App. D (Thm. D.1) |
| $\Lambda_s$ | Max. normalized centroid error at iteration $s$ | App. D (Thm. D.1) |
| $\alpha_{\max}$ | Max. anomaly-effect norm | App. D (Thm. D.1) |
| $\mathcal{A}_0$ | True anomalous subject set | Sec. 3 |
| $\rho_{im}^k$ | Cluster-$k$ mFPCA score, component $m$ | App. E |
| $\mathcal{L}_m^k$ | Null distribution of $\rho_{im}^k$ in cluster $k$ | App. E (Thm. E.1) |
| $\tilde{\rho}_m^k$ | Bootstrap draw from $\mathcal{L}_m^k$ | App. E (Thm. E.1) |
| $\mathcal{A}_0^{k,\epsilon}$ | Screened candidates in cluster $k$ at level $\epsilon$ | App. E (Thm. E.1) |
| $I_s, G_s$ | Screening split and potential outliers set | Alg. 7 |
| $p_{im}^{\text{emp}}$ | Empirical p-value for subject $i$, component $m$ | Alg. 8 |
| $\mathcal{W}_i$ | Anomalous time-window set for subject $i$ | Alg. 10 |
| $\bar{\boldsymbol{\mu}}^{(w)}$ | Cluster mean averaged over window $w$ | Alg. 9 |
| $\alpha/W$ | Window-wise Bonferroni threshold | Alg. 10 |